%% file: mainV4.tex
\documentclass[10pt,journal,compsoc]{IEEEtran}
\ifCLASSOPTIONcompsoc
  \usepackage[nocompress]{cite}
\else
  \usepackage{cite}
\fi

\ifCLASSINFOpdf
\else
\fi

\usepackage{amsmath}
\usepackage{amssymb}
\usepackage{graphicx}
\usepackage{booktabs}
\usepackage{algorithmic}
\usepackage{array}
\usepackage{stfloats}
\usepackage{url}
\usepackage{multirow, colortbl, booktabs}
\usepackage{overpic}
\usepackage{amsfonts}
\usepackage{bm}
\usepackage{relsize}
\usepackage{multirow}
\usepackage{threeparttable}
\usepackage{makecell}
\usepackage{bbm}
\usepackage{ragged2e} 
\usepackage{bbding}
\usepackage{colortbl}
\usepackage[dvipsnames]{xcolor}
\usepackage{caption}
\usepackage{subcaption}

\newcommand*{\belowrulesepcolor}[1]{%
  \noalign{%
    \kern-\belowrulesep 
    \begingroup 
      \color{#1}%
      \hrule height\belowrulesep 
    \endgroup 
    \vspace{-0.03mm}
  }%
} 
\newcommand*{\aboverulesepcolor}[1]{%
  \noalign{%
  \vspace{-0.03mm}
    \begingroup 
      \color{#1}%
      \hrule height\aboverulesep 
    \endgroup 
    \kern-\aboverulesep 
  }%
}

\usepackage{soul}
\definecolor{airforceblue}{rgb}{0.36, 0.54, 0.66}
\definecolor{skyblue}{rgb}{0.53, 0.81, 0.92}
\definecolor{frenchblue}{rgb}{0.0, 0.45, 0.73}

\definecolor{americanrose}{rgb}{1.0, 0.01, 0.24}
\definecolor{myred}{rgb}{0.753, 0.314, 0.275}
\definecolor{myblue}{rgb}{0.0, 0.24, 0.95}
\definecolor{tbl_gray}{gray}{0.85}
\newcommand{\RNum}[1]{\uppercase\expandafter{\romannumeral #1\relax}}

\newcommand{\ie}{\textit{i}.\textit{e}.}
\newcommand{\eg}{\textit{e}.\textit{g}.}

\newcommand\MYhyperrefoptions{bookmarks=true,bookmarksnumbered=true,
pdfpagemode={UseOutlines},plainpages=false,pdfpagelabels=true,
colorlinks=true,linkcolor={americanrose},citecolor={myblue},urlcolor={myblue}}
\usepackage[\MYhyperrefoptions,pdftex]{hyperref}

\usepackage[capitalize]{cleveref}
\crefname{section}{Sec.}{Sec.}

\newlength\savedwidth

\begin{document}

\title{A Survey on Open-Vocabulary Detection and Segmentation: Past, Present, and Future}

\author{Chaoyang Zhu, Long Chen

\IEEEcompsocitemizethanks{
\IEEEcompsocthanksitem Chaoyang Zhu and Long Chen are with the Department of Computer Science and Engineering, The Hong Kong University of Science and Technology, Kowloon, Hong Kong.\protect\\
E-mail: sean.zhuh@gmail.com, longchen@ust.hk.\protect\\
The corresponding author is Long Chen.
}

\thanks{Manuscript received July 18, 2023; revised April 15, 2024.}
}

\markboth{A SUBMISSION TO IEEE TRANSACTIONS ON PATTERN ANALYSIS AND MACHINE INTELLIGENCE}%
{Shell \MakeLowercase{\textit{et al.}}: Bare Demo of IEEEtran.cls for Computer Society Journals}

\IEEEtitleabstractindextext{

\justifying

\begin{abstract}

As the most fundamental scene understanding tasks, object detection and segmentation have made tremendous progress in deep learning era. Due to the expensive manual labeling cost, the annotated categories in existing datasets are often small-scale and pre-defined, \emph{i.e.}, state-of-the-art fully-supervised detectors and segmentors fail to generalize beyond the closed vocabulary. To resolve this limitation, in the last few years, the community has witnessed an increasing attention toward \textbf{Open-Vocabulary Detection (OVD)} and \textbf{Segmentation (OVS)}. By ``open-vocabulary'', we mean that the models can classify objects beyond pre-defined categories. In this survey, we provide a comprehensive review on recent developments of OVD and OVS. A taxonomy is first developed to organize different tasks and methodologies. We find that the permission and usage of weak supervision signals can well discriminate different methodologies, including: \emph{visual-semantic space mapping, novel visual feature synthesis, region-aware training, pseudo-labeling, knowledge distillation, and transfer learning}. The proposed taxonomy is universal across different tasks, covering object detection, semantic/instance/panoptic segmentation, 3D and video understanding. The main design principles, key challenges, development routes, methodology strengths, and weaknesses are thoroughly analyzed. In addition, we benchmark each task along with the vital components of each method in appendix and updated online at \href{https://github.com/seanzhuh/awesome-open-vocabulary-detection-and-segmentation}{awesome-ovd-ovs}. Finally, several promising directions are provided and discussed to stimulate future research.

\end{abstract}


}

\maketitle

\IEEEdisplaynontitleabstractindextext

\IEEEpeerreviewmaketitle


\IEEEraisesectionheading{
\section{Introduction}
\label{sec:intro}
}

\input{sections/V4/1.intro}


\section{Preliminaries}
\label{sec:preliminaries}

\input{sections/V4/2.preliminaries}


\section{Zero-Shot Detection (ZSD)}
\label{sec:zsd}

\input{sections/V4/3.zsd}


\section{Zero-Shot Segmentation (ZSS)}
\label{sec:zss}

\input{sections/V4/4.zss}


\section{Open-Vocabulary Detection (OVD)}
\label{sec:ovd}

\input{sections/V4/5.ovd}


\section{Open-Vocabulary Segmentation (OVS)}
\label{sec:ovs}

\input{sections/V4/6.ovs}


\section{Open-Vocabulary Beyond Images}
\label{sec:ov3d}

\input{sections/V4/7.ov3d}


\section{Challenges and Outlook}
\label{sec:challenges-outlook}

\input{sections/V4/8.challenges-outlook}


\section{Conclusion}
\label{sec:conclusion}

\input{sections/V4/9.conclusion}




\ifCLASSOPTIONcaptionsoff
  \newpage
\fi

{
\tiny
\bibliographystyle{IEEEtran}
\bibliography{refs}
}

\input{sections/V4/bio}

\clearpage
\input{sections/V4/supplement.tex}

\end{document}

%% file: sections/V4/1.intro.tex
\IEEEPARstart{O}{bject} detection and segmentation are core high-level scene perception tasks in computer vision. They are cornerstones of numerous real-world applications such as autonomous driving~\cite{nuscenes,bevformer}, intelligent robotics~\cite{embodied-navigation,vln}, to name a few. Given an image, video, or a set of point clouds, object detection~\cite{faster-rcnn,fpn,3detr} predicts tightly-enclosed bounding boxes along with class labels, while segmentation task groups pixels or points into a semantically coherent area or volume (semantic segmentation)~\cite{fcn,maskformer}, an instance with a distinctive ID (instance segmentation)~\cite{mask-rcnn}, or a combination of both (panoptic segmentation)~\cite{panoptic-segmentation}.

The past decade has witnessed a steady and tremendous progress in object detection and segmentation tasks brought by CNN-based and Transformer-based models~\cite{faster-rcnn,mask-rcnn,focal,detr,maskformer,mask2former}. However, existing detectors/segmentors can only localize pre-defined semantic concepts (labels) within a specific dataset. They are typically at small-scale, \emph{e.g.}, 20 classes in Pascal VOC~\cite{pascal-voc}, 80 classes in COCO~\cite{mscoco}, even the largest LVIS~\cite{lvis} dataset merely annotates 1,203 categories. On the contrary, our human perception system can associate arbitrary visual concepts with open-ended class names or natural language descriptions. The closed-set limitation greatly hinders the utilization of current detectors/segmentors in real-world applications. 

To resolve the closed-vocabulary constraint for scene perception tasks, research endeavors have been devoted to \textbf{zero-shot} or \textbf{open-vocabulary} detection and segmentation:

\noindent\textbf{Zero-Shot.} In the early stage, zero-shot detection (ZSD)~\cite{sb-lab,zsd-accv,zsd-tcsvt} and segmentation (ZSS)~\cite{zs3net,zsi} are first proposed as an attempt \textit{w/o allowing the access to any unannotated unseen visual object}. Typical methods always replace the learnable classifier (a fully-connected layer) with fixed \textit{semantic embeddings}, \emph{e.g.}, Word2Vec~\cite{word2vec}, GloVe~\cite{glove}, or BERT~\cite{bert}. Semantic embeddigs can transfer knowledge from seen (base) categories to unseen (novel) ones. However, due to the unsupervised training on text corpus only, semantic embeddings lack the alignment with visual modality, \ie, they are noisy to serve as anchors for calibrating visual space and impede performance improvement~\cite{vild}. 

\noindent\textbf{Open-Vocabulary.} Open-vocabulary detection~\cite{ovr-cnn,vild} and segmentation~\cite{xpm,openseg,lseg} \textit{allow the model to train on images with unannotated novel objects}. The closed-set limitation is addressed via weak supervision signals, \emph{i.e.}, image-text pairs (image-caption pairs and image-level labels), or large pretrained Vision-Language Models (VLMs), such as CLIP~\cite{clip}. VLMs text encoder (VLMs-TE) encodes template prompts~\cite{clip} filled with class names into \emph{text embeddings}\footnote{Both text and semantic embeddings are label embeddings.}, which are deemed as the frozen classifier. They well align with image modality due to the image-text contrastive pretraining~\cite{clip}. Therefore, OVD and OVS achieve a huge performance leap compared to ZSD and ZSS. 

Given the great application value, a plethora of methods have been proposed in recent years, making it hard for researchers to keep pace with them. However, to the best of our knowledge, only a few related surveys are available, focusing on limited tasks and settings\footnote{In this survey, ``zero-shot'' and ``open-vocabulary'' are regarded as two different settings following prior work.}. We include zero-shot setting as a complement to open-vocabulary setting for two reasons: 1) both settings aim for resolving the closed-vocabulary constraint; 2) methodologies under the two settings are interchangeable, \emph{e.g.}, the novel visual feature synthesis methodology (discussed later) under zero-shot setting can be transferred to open-vocabulary setting with negligible effort. The motivation for separating detection and segmentation task is that their definitions, training losses, architectures, evaluation metrics, and datasets are different. These tasks are also advanced separately over the past decade in literature, we discuss on unifying open-vocabulary detection and segmentation in~\cref{sec:challenges-outlook}.

\begin{figure*}[t!]
    \centering
    \begin{overpic}[width=18cm, height=10.35cm]{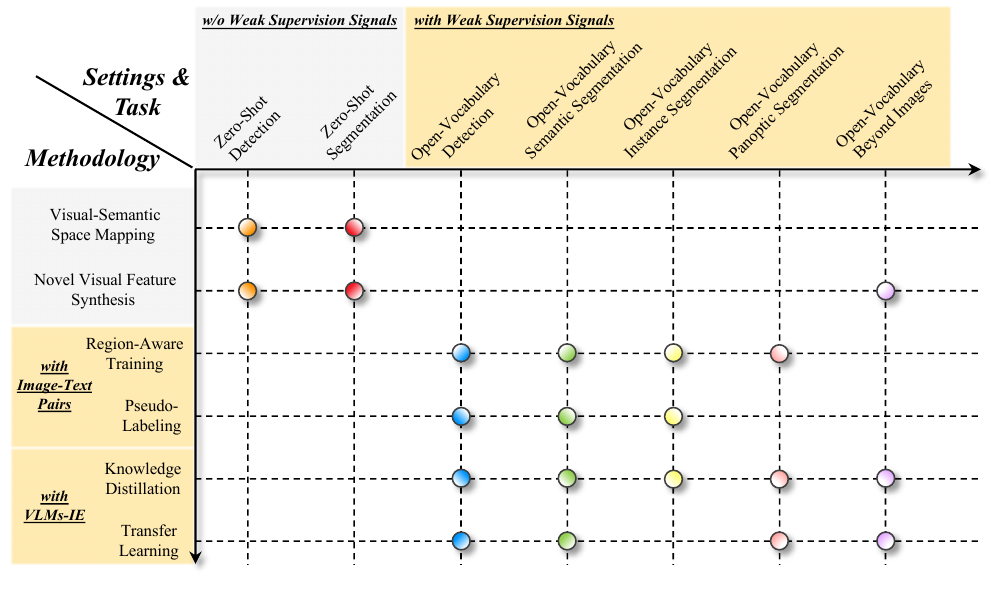}
        \put(26, 39){\scriptsize{LAB~\cite{sb-lab}}}
        \put(26, 37.5){\scriptsize{PL~\cite{polarity}}}
        \put(26, 36){\scriptsize{MS-Zero~\cite{mszero++}}}
        \put(26, 33.25){\scriptsize{DELO~\cite{delo}}}
        \put(26, 31.75){\scriptsize{GTNet~\cite{gtn}}}
        \put(26, 30.25){\scriptsize{RRFS~\cite{rrfs}}}
        \put(37, 39){\scriptsize{SPNet~\cite{spnet}}}
        \put(37, 37.5){\scriptsize{JoEm~\cite{joem}}}
        \put(37, 36){\scriptsize{PMOSR~\cite{pmosr}}}
        \put(37, 32.5){\scriptsize{ZS3Net~\cite{zs3net}}}
        \put(37, 31){\scriptsize{CaGNet~\cite{cagnet}}}
        \put(46.5, 27.5){\scriptsize{OVR-CNN~\cite{ovr-cnn}}}
        \put(46.5, 26.2){\scriptsize{MDETR~\cite{mdetr}}}
        \put(46.5, 24.9){\scriptsize{VLDet~\cite{vldet}}}
        \put(46.5, 23.6){\scriptsize{OV-DETR~\cite{ov-detr}}}
        \put(46.5, 21){\scriptsize{GLIP~\cite{glip}}}
        \put(46.5, 19.5){\scriptsize{PB-OVD~\cite{pb-ovd}}}
        \put(46.5, 18){\scriptsize{Detic~\cite{detic}}}
        \put(46.5, 14.9){\scriptsize{ViLD~\cite{vild}}}
        \put(46.5, 13.4){\scriptsize{DetPro~\cite{detpro}}}
        \put(46.5, 11.9){\scriptsize{BARON~\cite{baron}}}
        \put(46.5, 7){\scriptsize{F-VLM~\cite{f-vlm}}}
        \put(57.5, 26){\scriptsize{GroupViT~\cite{group-vit}}}
        \put(57.5, 24.5){\scriptsize{OpenSeg~\cite{openseg}}}
        \put(57.5, 19.5){\scriptsize{TTD~\cite{ttd}}}
        \put(57.5, 14){\scriptsize{GKC~\cite{gkc}}}
        \put(57.5, 12.5){\scriptsize{SAM-CLIP~\cite{sam-clip}}}
        \put(57.5, 9.4){\scriptsize{LSeg~\cite{lseg}}}
        \put(57.5, 7.9){\scriptsize{ZegFormer~\cite{zegformer}}}
        \put(57.5, 6.5){\scriptsize{OVSeg~\cite{ovseg}}}
        \put(57.5, 5.1){\scriptsize{SAN~\cite{san}}}
        \put(68.5, 25){\scriptsize{CGG~\cite{cgg}}}
        \put(68.5, 19){\scriptsize{XPM~\cite{xpm}}}
        \put(68.5, 13){\scriptsize{OV-SAM~\cite{ov-sam}}}
        \put(79, 25){\scriptsize{APE~\cite{ape}}}
        \put(79, 13){\scriptsize{PADing~\cite{pading}}}
        \put(79, 8){\scriptsize{FC-CLIP~\cite{fc-clip}}}
        \put(79, 6.5){\scriptsize{ODISE~\cite{odise}}}
        \put(89, 32.5){\scriptsize{3DGenZ~\cite{3dgenz}}}
        \put(89, 31){\scriptsize{SeCondPoint~\cite{secondpoint}}}
        \put(89, 20){\scriptsize{OV-3DET~\cite{ov-3det}}}
        \put(89, 18.5){\scriptsize{FM-OV3D~\cite{fm-ov3d}}}
        \put(89, 14){\scriptsize{CoDA~\cite{coda}}}
        \put(89, 12.5){\scriptsize{OpenScene~\cite{openscene}}}
        \put(89, 7){\scriptsize{OpenVIS~\cite{openvis}}}
    \end{overpic}
    \vspace{-15pt}
    \caption{The proposed taxonomy. Typical models are shown in each category. VLMs-IE denote the image encoder of VLMs.}
    \label{fig:taxonomy}
\end{figure*}

\begin{figure}[t!]
\centering
\begin{subfigure}{\linewidth}
     \includegraphics[width=8.9cm]{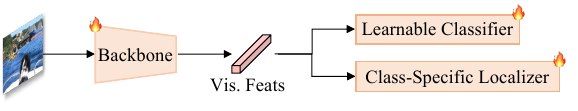}
     \vspace{-15pt}
     \caption{Canonical Detectors/Segmentors}
     \label{subfig:comp-1}
\end{subfigure}
\begin{subfigure}{\linewidth}
     \vspace{+5pt}
     \includegraphics[width=8.9cm]{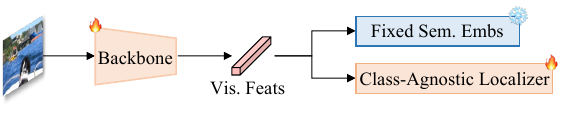}
     \vspace{-15pt}
     \caption{Visual-Semantic Space Mapping}
     \label{subfig:comp-2}
\end{subfigure}
\begin{subfigure}{\linewidth}
     \vspace{+5pt}
     \includegraphics[width=8.9cm]{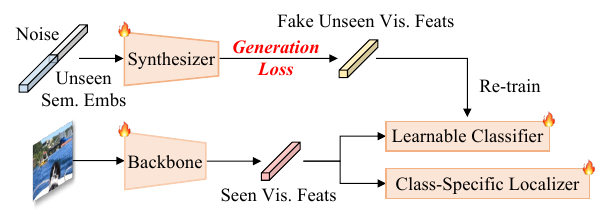}
     \vspace{-15pt}
     \caption{Novel Visual Feature Synthesis}
     \label{subfig:comp-3}
\end{subfigure}
\begin{subfigure}{\linewidth}
     \vspace{+5pt}
     \includegraphics[width=8.9cm]{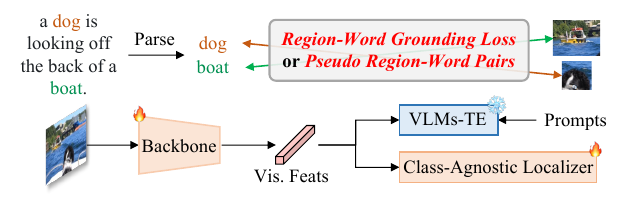}
     \vspace{-15pt}
     \caption{Region-Aware Training and Pseudo-Labeling}
     \label{subfig:comp-4}
\end{subfigure}
\begin{subfigure}{\linewidth}
     \vspace{+5pt}
     \includegraphics[width=8.9cm]{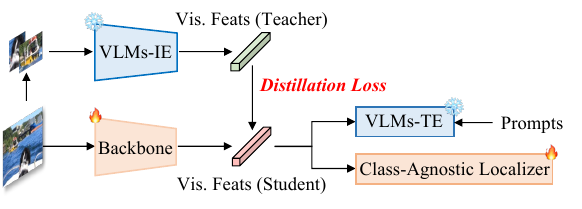}
     \vspace{-15pt}
     \caption{Knowledge Distillation}
     \label{subfig:comp-5}
\end{subfigure}
\begin{subfigure}{\linewidth}
     \vspace{+5pt}
     \includegraphics[width=8.9cm]{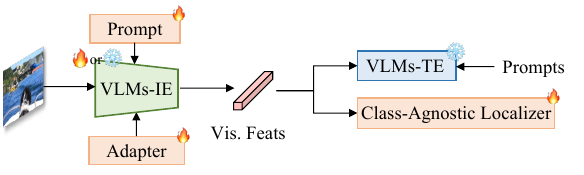}
     \vspace{-15pt}
     \caption{Transfer Learning}
     \label{subfig:comp-6}
\end{subfigure}
\caption{A general comparison of each methodology. ``Vis. Feats'' and ``Sem. Embs'' are visual features and semantic embeddings~\cite{word2vec,glove,fasttext,bert}, respectively.}
\label{fig:methodology-comparison}
\vspace{-10pt}
\end{figure}

In this paper, we provide a comprehensive review on different scene understanding tasks and settings including zero-shot/open-vocabulary detection, zero-shot/open-vocabulary semantic/instance/panoptic segmentation, as well as 3D scene and video understanding. To organize methods from these diverse tasks and settings, we need to answer the question: \emph{How to build a taxonomy that differentiates zero-shot and open-vocabulary settings while in the meantime abstracts universal methodologies across tasks?} We find that, \emph{whether or not to permit access to weak supervision signals, and if permitted, which one of them to utilize is key to categorization}. As shown in~\cref{fig:taxonomy}, zero-shot and open-vocabulary settings are differentiated by the permission of weak supervision signals, and different methodologies differ on which weak supervision signal to use during training. Under each setting, different tasks can share the same taxonomy. 

Concretely, \textbf{ZSD} and \textbf{ZSS} are not allowed to access weak supervision signals. To generalize beyond seen objects, besides substituting the learnable classifier in closed-set detectors/segmentors with fixed semantic embeddings, the class-specific localizer is also switched to a class-agnostic one, \emph{i.e.}, the output dimension of last regression layer is four ([$x_1,y_1,x_2,y_2$] or [$x,y,w,h$]) instead of four times the number of test classes (see~\cref{subfig:comp-1,subfig:comp-2}). Methodologies under zero-shot setting can be grouped into:

\noindent \underline{\textbf{Visual-Semantic Space Mapping.}} Though the visual and semantic space may bear discriminative capabilities in one modality, there is no direct cross-modality training mechanisms mining mutual relationships between both spaces. Thus, learning a mapping from visual to semantic space, semantic to visual space, or a joint mapping of visual-semantic space via tailored losses is crucial to enable such a reliable cross-space similarity measurement. However, due to the lack of unseen annotations, the prediction confidence is always biased toward seen classes.

\noindent \underline{\textbf{Novel Visual Feature Synthesis.}} This methodology utilizes an additional generative model~\cite{gmmn,gan,cvae} to synthesize fake unseen visual features conditioned on semantic embeddings and random noises, which transfer the problem into a ``fully-supervised'' setting. The generation loss in~\cref{subfig:comp-3} is to approximate the underlying distribution of real visual features. Then, the classifier embedded in the detector head is retrained on both pristine real seen and generated unseen visual features. Since non-seen regions are typically classified as background, this methodology alleviates both the confusion between novel and background concepts, and the bias issue in previous methodology. A more detailed pipeline is given in~\cref{fig:novel-visual-feature-synthesis-flowchart}.

Once allowed to access the weak supervision signals, \textbf{OVD} and \textbf{OVS} methodologies can be mainly categorized into four types: \textbf{Region-Aware Training} mainly leverages image-text pairs \emph{w/o} VLMs image encoder (VLMs-IE); \textbf{Pseudo-Labeling} leverages both image-text pairs and VLMs-IE (optional); \textbf{Knowledge Distillation} and \textbf{Transfer Learning} mainly utilize VLMs-IE and seldomly trained on image-text pairs. Note that open-vocabulary setting also adopt a class-agnostic localization branch, and all four methodologies use VLMs-TE to encode class names into frozen text embeddings as the classifier. We now detail the four methodologies as the following.

\noindent \underline{\textbf{Region-Aware Training.}} As~\cref{subfig:comp-4} shows, regions and words are ``\emph{implicitly}'' aligned on the cheap and abundant image-text pairs besides ground-truth annotations. Its main characteristic is the imposed bi-directional weakly-supervised grounding or contrastive loss~\cite{ovr-cnn} (see~\cref{eq:i2t,eq:t2i}). The loss pulls regions and words within the same image-text pair close while pushing away other negatives inside a batch. However, the exact one-to-one correspondence between regions and words remains unknown due to its weakly supervised nature. Nonetheless, during training the model encounters a larger vocabulary containing novel classes, hence improving the base-to-novel generalization ability. Models leveraging bipartite matching~\cite{detr} and region-level grounding annotations~\cite{vg,refcoco/+,refcocog} are also grouped into this methodology.

\noindent \underline{\textbf{Pseudo-Labeling.}} In contrast, pseudo-labeling knows the exact correspondence between words and regions. It ``\emph{explicitly}'' constructs either pseudo region-word or region-caption pairs for novel classes in a teacher-student framework. It can be seen as a ``\textbf{hard}'' alignment, \emph{i.e.}, one region can only correspond to one word, and vice versa. In contrast, region-aware training is ``\textbf{soft}'' alignment, where one word/region may correspond to multiple regions/words weighted by \emph{softmax}. Pesudo-labeling can adopt VLMs-IE as teachers to produce pseudo labels, it can also be deemed as self-training w/o VLMs-IE (the teacher is the model itself). One defect of pseudo-labeling is that it requires knowing novel class names in advance during training which is impractical and breaks open-vocabulary setting.

\noindent \underline{\textbf{Knowledge Distillation.}} VLMs (\eg, CLIP) trained via contrastive learning yield superior zero-shot recognition ability across various downstream tasks. Shown in~\cref{subfig:comp-5}, this methodology mainly distills region embeddings from VLMs-IE into a detector using only downstream detection/segmentation data. Since region-of-interests (RoIs) may contain novel objects, mimicking region embeddings of the teacher enables the detector to approach CLIP visual-semantic space more effectively. A more detailed framework is given in~\cref{fig:knowledge-distillation-framework}.

\noindent \underline{\textbf{Transfer Learning.}} We term the usage of transferring a pretrained VLMs-IE to downstream perception tasks as transfer learning. In knowledge distillation, both the detector backbone and VLMs-IE are employed during training (increasing computational overhead and memory footprint) and the VLMs-IE is typically discarded in inference. In contrast, transfer learning utilizes only VLMs-IE in both training and test stages, adding negligible computation cost. Shown in~\cref{subfig:comp-6} and~\cref{fig:transfer-learning-based}, these methods can be further categorized into: 1) frozen VLMs-IE as feature extractor; 2) fine-tuning VLMs-IE on downstream data; 3) learning visual prompts~\cite{vpt} or a lightweight adapter~\cite{tip-adapter,vl-adapter,san} to the frozen VLMs-IE in a parameter-efficient way.


The above taxonomy can also be seamlessly applied to both open-vocabulary 3D scene and video understanding. Due to the lack of large-scale point clounds-text or video-text pairs, large VLMs in the image domain are exploited as an intermediate bridge between point clouds and texts or between videos and texts. For instance, pseudo-labels generated by OVD detectors in the 2D domain are associated with semantic labels via VLMs and they are back projected to 3D space as supervision.

The remainder of the paper is organized as follows: ~\cref{sec:preliminaries} introduces preliminary background. Then we review ZSD and ZSS in~\cref{sec:zsd} and ~\cref{sec:zss}, OVD and OVS in~\cref{sec:ovd} and~\cref{sec:ovs}, respectively. Open-vocabulary 3D scene and video understanding are covered in~\cref{sec:ov3d}. We point out challenges and outlook in~\cref{sec:challenges-outlook} and conclude in~\cref{sec:conclusion}. Additional benchmarks with vital components of each method are listed in appendix.

%% file: sections/V4/2.preliminaries.tex
 \subsection{Problem Definition}
\label{subsec:problem-definition}

\begin{table}[t]
\setlength{\tabcolsep}{3pt}
    \centering
    \captionsetup{labelsep=colon}
    \caption{Differences of ZSD/ZSS and OVD/OVS. G and Non-G denote generalized and non-generalized evaluation.}
    \label{tab:zs-ov-diff}
    \vspace{-5pt}
    \resizebox{\linewidth}{!}{
    \begin{tabular}{cccccc}
    \toprule
        \multirow{5}{*}{External Information} & \multicolumn{4}{c}{ZSD/ZSS} & \multirow{5}{*}{OVD/OVS} \\
    \noalign{\smallskip}
    \cline{2-5}
    \noalign{\smallskip}
         & \multicolumn{2}{c}{Inductive} & \multicolumn{2}{c}{Transductive} & \\
    \noalign{\smallskip}
    \cline{2-5}
    \noalign{\smallskip}
         & G & Non-G & G & Non-G & \\
    \midrule
        \makecell[c]{Occurrence of \\ Unlabeled \& Novel \\ Objects} & {\color{BrickRed} \XSolidBrush} & {\color{BrickRed} \XSolidBrush} & {\color{ForestGreen} \Checkmark} & {\color{ForestGreen} \Checkmark} & {\color{ForestGreen} \Checkmark} \\
    \midrule
        Image-Text Pairs & {\color{BrickRed} \XSolidBrush} & {\color{BrickRed} \XSolidBrush} & {\color{BrickRed} \XSolidBrush} & {\color{BrickRed} \XSolidBrush} & {\color{ForestGreen} \Checkmark} \\
    \midrule
        Large VLMs & {\color{BrickRed} \XSolidBrush} & {\color{BrickRed} \XSolidBrush} & {\color{BrickRed} \XSolidBrush} & {\color{BrickRed} \XSolidBrush} & {\color{ForestGreen} \Checkmark} \\
    \midrule
        Test Classes & $\mathcal{C}_{B} \cup \mathcal{C}_{N}$ & $\mathcal{C}_{N}$ & $\mathcal{C}_{B} \cup \mathcal{C}_{N}$ & $\mathcal{C}_{N}$ & $\mathcal{C}_{B} \cup \mathcal{C}_{N}$ \\
    \bottomrule
    \end{tabular}}
\end{table}

The goal of OVD/OVS is to detect or segment unseen or novel classes that occupy semantically coherent regions or volumes within an image, video, or a set of point clouds. During its early stage of development, \textbf{inductive} ZSD/ZSS is first formulated to achieve this goal. ZSD/ZSS imposes a constraint that training images do not contain any unseen instance even if it is not annotated. To resolve the limitation, \textbf{transductive} ZSD~\cite{tlzsd} considers unannotated and unseen samples during training. However, inductive ZSD/ZSS has been the mainstream and more actively studied than transductive ZSD/ZSS. OVD/OVS can be deemed as a variant of transductive ZSD/ZSS that further allows the usage of weak supervision signals. Nonetheless, both zero-shot and open-vocabulary settings avoid annotated novel objects appearing in the training set. OVD/OVS splits the labeled set $\mathcal{C}$ of annotations into two disjoint subsets of base and novel categories. We denote them by $\mathcal{C}_B$ and $\mathcal{C}_N$, respectively. Note that $\mathcal{C}_B \cap \mathcal{C}_N=\varnothing$ and $\mathcal{C}=\mathcal{C}_B \cup \mathcal{C}_N$. Thus the labeled set for training and test is $\mathcal{C}_{train}=\mathcal{C}_{B}$ and $\mathcal{C}_{test}=\mathcal{C}_{B}\cup \mathcal{C}_{N}$. With this definition, the difference with closed-set tasks is clear, where $\mathcal{C}_{test}=\mathcal{C}_{train}=\mathcal{C}$. A complete comparison between ZSD/ZSS and OVD/OVS is listed in~\cref{tab:zs-ov-diff}.

\subsection{Related Domains and Tasks}
\label{subsec:related-domains}

In this subsection, we describe several highly related domains with OVD/OVS and summarize their differences. 

\noindent \textbf{Visual Grounding.} It grounds semantic concepts to image regions~\cite{refcoco/+,refcocog,referitgame,flickr30k,seqtr}. Specifically, the task can be divided into: 1) phrase localization~\cite{flickr30k} that grounds all nouns in the sentence; 2) referring expression comprehension~\cite{refcoco/+,refcocog,referitgame} that only grounds the referent in the sentence. In the latter, the referent is labeled not with a class name but with freeform natural language describing instance attributes, positions, and relationships with other objects or backgrounds. This task greatly expands the concepts that the model can ground but the vocabulary is still closed-set. An ideal way to achieve OVD and OVS is to scale the small-scale grounding datasets to web-scale datasets. However, the laborious labeling cost is non-negligible. 

\noindent \textbf{Weakly-Supervised Detection and Segmentation.} Without any bounding box or dense mask annotation, training labels of weakly-supervised setting~\cite{wsod-survey} only comprise image-level class names. The image-level labels indicate object classes that appear in the image. Many weakly-supervised learning techniques like multiple instance learning~\cite{mil} and weakly-supervised grounding loss have been introduced into OVD/OVS. However, weakly-supervised object detection and segmentation work under the closed-set setting.

\noindent \textbf{Open-Set Detection and Segmentation.} Open-set detection~\cite{overlooked,dropout-osd} and segmentation~\cite{oss1,oss2} stem from open-set recognition~\cite{osr1,osr2}. It requires classifying known classes and identifying a single ``\emph{unknown}'' class without further classifying exact classes. It is equivalent to setting $\mathcal{C}_{train}=\mathcal{C}_{B}$ and $\mathcal{C}_{test}=\mathcal{C}_{B}\cup \{u\}$ where $u$ represents the single ``\emph{unknown}'' class. The main target is to reject unknown classes that emerge unexpectedly and may hamper the robustness of the recognition system.

\noindent \textbf{Open-World Detection and Segmentation.} Open-world detection~\cite{ore,ow-detr} and segmentation~\cite{dmlnet} take a step further to open-set detection and segmentation. At time step $t$, objects are classified as $\mathcal{C}_B^t=\{c_1,c_2,\dots,c_k\}\cup \{u\}$. Then, unknown instances are labeled as newly known classes $\{c_{k+1},\dots,c_{k+m}\}$ by an oracle and added back to $\mathcal{C}_B^t$. At time $t$+1, detector is required to detect $\mathcal{C}_B^{t+1}=\{c_1,\dots,c_k,\dots,c_{k+m}\}\cup\{u\}$. After each time step, the number of unknown classes belonging to ``\emph{unknown}'' will decrease. This continual learning cycle repeats over the lifetime of the detector. The task aims at incrementally learning new classes without forgetting previously learned classes in a dynamic world. Note that the detector is not re-trained from scratch across time steps.

\noindent \textbf{Out-of-Distribution Detection.} In out-of-distribution detection~\cite{ood-energy,ood-generalized}, test samples are \emph{not} assumed to be drawn from the same distribution of training data (\ie, \emph{i.i.d.}). Specifically, distribution shift can be divided into: 1) \textbf{Semantic shift}, where out-of-distribution samples are from different classes. 2) \textbf{Covariate shift}, where out-of-distribution samples are from different domains but with same classes, such as sketches or adversarial examples. Out-of-distribution detection primarily focuses on the former which resembles open-set recognition. Nonetheless, it still does not require sub-classifying the single ``\emph{out-of-distribution}'' class which is the same as the ``\emph{unknown}'' class in open-set/world detection and segmentation.

\subsection{Canonical Closed-Set Detectors and Segmentors}
\label{subsec:canonical-detectors-segmentors}

\noindent\textbf{Object Detection.} Faster R-CNN~\cite{faster-rcnn} is a representative \textbf{two-stage} detector. Based on anchor boxes, the region proposal network (RPN) first hypothesizes potential object regions to separate foreground and background proposals by measuring their objectness scores. Then, an RCNN-style~\cite{fast-r-cnn} detection head predicts per-class probability and refines the locations of positive proposals. Meanwhile, \textbf{one-stage} detectors directly refine the positions of anchors without proposal generation stage. FCOS~\cite{fcos} regards each feature map grid within the ground-truth box as a positive anchor point and regresses its distances to the four edges of the target box. The centerness suppresses low-quality predictions of anchor points that are near the boundary of ground-truths. With the development of Transformers in NLP, Transformer-based detectors have dominated the literature recently. DETR~\cite{detr} reformulates object detection as a set matching problem with a Transformer encoder-decoder architecture. The learnable object queries attend to encoder output via cross-attention and specialize in detecting objects with different positions and sizes. Deformable DETR~\cite{deformable-detr} designs a multi-scale deformable attention mechanism that sparsely attends sampled points around queries to accelerate convergence. 

\noindent\textbf{Segmentation.} DeepLab~\cite{deeplab,deeplabv3+} enhances FCN~\cite{fcn} with dilated convolution, conditional random field, and atrous spatial pyramid pooling. Mask R-CNN~\cite{mask-rcnn} adds a parallel mask branch to Faster R-CNN and proposes RoI Align for instance segmentation. Following DETR, MaskFormer~\cite{maskformer} obtains mask embeddings from object queries, and performs dot-product with up-sampled pixel embeddings to produce segmentation maps. It transforms the per-pixel classification paradigm into a mask region classification framework. Mask2Former~\cite{mask2former} follows the same meta-architecture of MaskFormer and introduces a masked cross-attention module that only attends to predicted mask regions.

\subsection{Large Vision-Language Models (VLMs)}
\label{subsec:large-VLMs}

Large VLMs have demonstrated a superior zero-shot transfer capability benefited from large-scale pretraining, and found various applications in OVD/OVS. In this subsection, we review large VLMs and related fine-tuning techniques.

\noindent \textbf{Image-Text Contrastive Pretraining.} CLIP~\cite{clip} makes the first breakthrough via contrastively pretraining on 400M image-caption pairs. The pretraining objective simply aligns positive image-caption pairs within a batch, making it efficient and scalable to learn transferable representations. During inference, template prompts filled with class names (\eg, ``a photo of a [CLASS]'') are fed into the CLIP text encoder, and the output of a special token [EOS] is regarded as the text embedding. For image embedding, a multi-head self-attention pooling layer aggregates patch embeddings into a holistic representation. Both text and image embeddings are $l_2$ normalized to compute their pair-wise cosine similarity. By this means, text embeddings are regarded as a frozen classifier. Later, ALIGN~\cite{align} leverages one billion noisy image alt-text pairs for pretraining with the same architecture and contrastive loss in CLIP. 

\noindent \textbf{Self-Supervised Learning with Local Semantics.} The work of DINO~\cite{dino,dinov2} self-distills its knowledge on unlabeled images. Two different random transformations of the same input image are passed to the student and teacher network separately. The student receives both local and global views, while the teacher only receives global views of the image. By matching the predicted distributions of teacher and student, DINO encourages a local-to-global correspondence establishment. Different heads in the last self-attention block of DINO clearly separates different object boundaries, enabling unsupervised object localization and pseudo-labeling in OVD/OVS. Meanwhile, MAE~\cite{mae} designs an asymmetric encoder-decoder framework, the encoder only perceives non-masked image patches and the decoder regresses the pixel value of masked patches. Since CLIP behaves like bag-of-words~\cite{bow}, \emph{i.e.}, a bag of local objects matches a bag of semantic concepts without differentiating distinct local regions, MAE and DINO are mainly utilized to enhance local image feature representations for OVD/OVS.

\noindent \textbf{Text-to-Image (T2I) Diffusion Models.} T2I diffusion models~\cite{ddpm,stable-diffusion} are also trained on internet-scale data. The step-by-step de-noising process gradually evolves pure noise tensors into realistic images conditioned on languages. Clustering on its internal feature representations clearly separates different objects, which is much less noisy than CLIP~\cite{clip,clip-surgery,clipself}. Since to generate distinct objects the model has to differentiate semantic concepts, text-guided generation objective naturally imposes precise region-word correspondence learning for the model during pretraining. In contrast, CLIP may cheat by only performing bag-of-words classification~\cite{bow}.

\noindent \textbf{Parameter-Efficient Fine-Tuning (PEFT).} Given the computational cost and memory footprint of these large VLMs, current endeavors only fine-tune a small set of additional parameters, such as prompt tuning~\cite{coop,vpt}, adapters~\cite{vl-adapter,tip-adapter,san}. Compared to fully fine-tuning the whole model on downstream data, PEFT balances the trade-off between overfitting on downstream data and preserving the prior knowledge of VLMs.

%% file: sections/V4/3.zsd.tex
Removing training images containing unannotated unseen objects induces the main challenge for ZSD. Hence, models resort to transfer knowledge of semantic embeddings~\cite{word2vec,glove,fasttext,bert} that are unsupervisedly trained on text corpus alone from seen to unseen classes. Methods in this section can be discriminative or generative: 1) visual-semantic space mapping is discriminative that seeks to maximize separation between the decision boundaries of ambiguous classes in visual, semantic, or common embedding space, especially the background and unseen classes; 2) novel visual feature synthesis is generative that uses a synthesizer for generating unseen visual features to bridge the data scarcity gap between seen and unseen classes.


\subsection{Visual-Semantic Space Mapping}
\label{subsec:zsd-vssm}

\subsubsection{Learning a Mapping from Visual to Semantic Space}
\label{subsubsec:zsd-vssm-v2s}

This mapping assumes the intrinsic structure of the semantic space is discriminative and it can well reflect the inter-class relationships. Besides the two modifications made to canonical detectors in~\cref{subfig:comp-2}, a linear layer (mapping function) is additionally added to the backbone to make the dimension of visual features the same as semantic embeddings.


\noindent\textbf{Two-Stage Methods.} At the time when ZSD is proposed, two-stage detectors~\cite{faster-rcnn} dominate closed-set object detection task. ZSD is first proposed by Bansal \emph{et al.}~\cite{sb-lab}. RoI features in Faster-RCNN are linearly projected into semantic space driven by a max-margin loss, then classified by GloVe embeddings~\cite{glove}. Bansal \emph{et al.} propose two techniques to remedy the ambiguity between background and unseen concepts. One is a fixed label vector ([1,0,...,0]) for modeling background, which is hard to cope with the high background visual variances. The other is dynamically assigning multiple classes from WordNet~\cite{wordnet} belonging to neither seen nor unseen classes to the background. A contemporaneous work~\cite{zsd-accv,zsd-ijcv} proposes a meta-class clustering loss besides the max-margin separation loss. It groups similar concepts to improve the separation between semantically-dissimilar concepts and reduce the noise in word vectors. Luo \emph{et al.}~\cite{ca-zsd} provide external relationship knowledge graph as pairwise potentials besides unary potentials in the conditional random field to achieve context awareness. ZSDTD~\cite{zsdtd} leverages textual descriptions to guide the mapping process. The textual descriptions are a general source for improving ZSD due to its rich and diverse context compared to a single word vector. Following polarity loss~\cite{polarity}, BLC~\cite{blc} develops a cascade architecture to progressively learn the mapping with an external vocabulary and a background learnable RPN to model background appropriately. Rahman \emph{et al.}~\cite{tlzsd} explore transductive generalized ZSD via fixed and dynamic pseudo-labeling strategies to promote training in unseen samples. Recently SSB~\cite{ssb} establishes a simple but strong baseline. It carefully ablates model characteristics, learning dynamics, and inference procedures from a myriad of design options. 


\noindent\textbf{One-Stage Methods.} Besides two-stage methods, applying one-stage detectors~\cite{yolov2,yolov3,focal} to ZSD is also explored. To improve the low recall rate of novel objects, ZS-YOLO~\cite{zsd-tcsvt} conditions the objectness branch of YOLO~\cite{yolov3} on the combination of semantic attributes, visual features, and localization output instead of visual features alone. HRE~\cite{hre} constructs two parallel visual-to-semantic mapping branches for classification. One is a convex combination of semantic embeddings, while the other maps grid features associated with positive anchors into the semantic space. Later Rahman \emph{et al.}~\cite{polarity} design a polarity loss that explicitly maximizes the margin between predictions of target and negative classes based on focal loss~\cite{focal}. A vocabulary metric learning approach is also proposed to provide a richer and more complete semantic space for learning the mapping. Similar to the hybrid branches in HRE~\cite{hre}, Li \emph{et al.}~\cite{cg-zsd} perform the prediction of super-classes and fine-grained classes in parallel. 

\subsubsection{Learning a Joint Mapping of Visual-Semantic Space}
\label{subsubsec:zsd-vssm-jvs}


Learning a mapping from visual to semantic space neglects the discriminative structure of visual space itself. MS-Zero~\cite{mszero++} demonstrates that classes can have poor separation in semantic space but are well separated in visual space, and vice versa. Hence it exploits this complementary information via two unidirectional mapping functions. Similarity metrics are calculated in both spaces which are then averaged as the final prediction. Similar to previous works~\cite{hre,cg-zsd}, DPIF~\cite{dpif} proposes a dual-path inference fusion module. It integrates empirical analysis of unseen classes by analogy with seen classes (past knowledge) into the basic knowledge transfer branch. The association predictor learns unseen concepts using training data from a group of associative seen classes as their pseudo instances. ContrastZSD~\cite{contrast-zsd} proposes to contrast seen region features to make the visual space more discriminative. It contrasts seen region features with both seen and unseen semantic embeddings under the guidance of semantic relation matrix.

\subsubsection{Learning a Mapping from Semantic to Visual Space}
\label{subsubsec:zsd-vssm-s2v}


Zhang \emph{et al.}~\cite{s2v} argue that learning a mapping from visual to semantic space or a joint space will shrink the variance of projected visual features and thus aggravates the hubness problem~\cite{hubness}, \emph{i.e.}, the high-dimensional visual features are likely to be embedded into a low dimensional area of incorrect labels. Hence they embed semantic embeddings to the visual space via a least square loss.

\subsection{Novel Visual Feature Synthesis}
\label{subsec:zsd-nvfs}

\begin{figure}[t]
    \centering    \includegraphics[width=9.1cm,height=4.36cm]{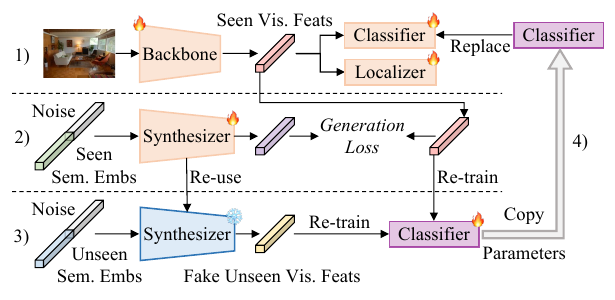}
    \vspace{-2em}
    \caption{Flowchart of novel visual feature synthesis.}
    \label{fig:novel-visual-feature-synthesis-flowchart}
\vspace{-1ex}
\end{figure}

\noindent\textbf{Overview.} To enable recognition of novel concepts, novel visual feature synthesis produces \emph{fake} unseen visual features as training samples for a new classifier. This methodology follows a multi-stage pipeline (as shown in~\cref{fig:novel-visual-feature-synthesis-flowchart}): 1) Train the base model with seen classes annotations in a fully-supervised manner. 2) Train the feature synthesizer $G: \mathcal{W} \times \mathcal{Z} \mapsto \tilde{\mathcal{F}}$ on seen semantic embeddings $\mathbf{w} \in \mathcal{W}_s \in \mathbb{R}^{d}$ and real seen visual features $\mathbf{f}_s \in \mathcal{F}_s \in \mathbb{R}^c$ extracted from the base model to learn the underlying distribution of visual features. 3) Conditioned on the unseen semantic embeddings $\mathbf{w} \in \mathcal{W}_u$ and a random noise vector $\mathbf{z} \sim \mathcal{N}(0,1)$, the synthesizer generates novel unseen visual features. A new classifier is retrained on fake unseen and real seen visual features, while the remaining parts of the base model are kept frozen. 4) Finally, the new classifier is plugged back into the base model. Note that the noise vector perturbs the synthesizer to produce various visually diverging features given the semantic embeddings.


DELO~\cite{delo} leverages conditional variational auto-encoder~\cite{cvae} with three consistency losses forcing the generated visual features to be coherent with the original real ones on the predicted objectness score, category, and class semantic. Then, it~\cite{delo} retrains the objectness branch to assign high confidence scores on both seen and unseen objects. This is to mitigate the low recall rate of unseen objects. Later on, Hayat \emph{et al.}~\cite{su} use the same class consistency loss but they adopt the mode seeking regularization~\cite{mode-seeking} which maximizes the distances of generated data points \emph{w.r.t} their noise vectors. At the same time with DELO~\cite{delo}, GTNet~\cite{gtn} proposes an IoU-aware synthesizer based on wasserstein generative adversarial network~\cite{wgan}. Since DELO is only trained on ground-truths that encloses the object boundary tightly, so its synthesizer can not generate unseen RoI features with diverse spatial context clues. That is, the retrained classifier can not correctly classify RoIs that loosely encloses the object boundary. To mitigate this context gap between unseen RoI features from RPN and those synthesized by the generator, GTNet~\cite{gtn} randomly samples foreground and background RoIs as the additional generation target, thus making the new classifier robust to various degrees of context. RRFS~\cite{rrfs} proposes an intra-class semantic diverging loss and an inter-class structure preserving loss. The former pulls positive synthesized features lying within the hypersphere of the corresponding noise vector close while pushing away those generated from distinct noise vectors. The latter constructs a hybrid feature pool of real and fake features to avoid mixing up the inter-class relationship.

%% file: sections/V4/4.zss.tex
ZSS takes a step further than ZSD at a finer pixel-level granularity. We cover zero-shot semantic and instance segmentation in this section as a complement to ZSD.


\subsection{Zero-Shot Semantic Segmentation (ZSSS)}
\label{subsec:zss-zsss}

\subsubsection{Visual-Semantic Space Mapping}
\label{subsubsec:zss-zsss-vsm}


\textbf{Learning a Mapping from Visual to Semantic Space.} SPNet~\cite{spnet} is the first work that proposes the zero-shot semantic segmentation task. It directly maps pixel features into semantic space optimized by the canonical \emph{cross-entropy} loss. During inference, SPNet calibrates seen predictions by subtracting a factor tuned on a held-out validation set. 


\noindent \textbf{Learning a Joint Mapping of Visual-Semantic Space.} Hu \emph{et al.}~\cite{ul} argue that the noisy and irrelevant samples in seen classes have negative effects on learning visual-semantic correspondence. An uncertainty-aware loss is proposed to adaptively strengthen representative samples while an attenuating loss for uncertain samples with high variance estimation. JoEm~\cite{joem} learns a joint embedding space via the proposed boundary-aware regression loss and semantic consistency loss. At the test stage, semantic embeddings are transformed into semantic prototypes acting as a nearest-neighbor classifier (without the classifier retraining stage in~\cref{subsubsec:zss-zsss-nvfs}). The apollonius calibration inference technique is further proposed to alleviate the bias problem. 


\noindent \textbf{Learning a Mapping from Semantic to Visual Space.} Kato \emph{et al.}~\cite{vm} propose a variational mapping from semantic space to visual space via sampling the conditions (mimicking the support images in few-shot semantic segmentation~\cite{panet}) from the predicted distribution. PMOSR~\cite{pmosr} abstracts a set of seen visual prototypes, then trains a projection network mapping seen semantic embeddings to these prototypes. Similar to JoEm~\cite{joem}, since one can simply project unseen semantic embeddings to unseen prototypes for classification, new unseen classes can be flexibly added in inference without classifier retraining. An open-set rejection module is further proposed to prevent unseen classes from directly competing with seen classes.

\subsubsection{Novel Visual Feature Synthesis}
\label{subsubsec:zss-zsss-nvfs}


Concurrent with SPNet~\cite{spnet}, ZS3Net~\cite{zs3net} conditions the synthesizer~\cite{gmmn} on adjacency graph encoding structural object arrangement to capture contextual cues for the generation process. CSRL~\cite{csrl} transfers the relational structure constraint in the semantic space including point-wise, pair-wise, and list-wise granularities to the visual feature generation process. However, in both methods, the \textbf{mode collapse problem}, \emph{i.e.}, the generator often ignores the random noise vectors appended to the semantic embeddings and produces limited visual diversity, hindering the effectiveness of generative models. CaGNet~\cite{cagnet} addresses this problem by replacing the simple noise with contextual latent code, which captures pixel-wise contextual information via dilated convolution and adaptive weighting between different dilation rates. Following CaGNet~\cite{cagnet}, SIGN~\cite{sign} also substitutes the noise vector but with a spatial latent code incorporating the relative positional encoding. While previous ZS3Net~\cite{zs3net} and CaGNet~\cite{cagnet} simply discard pseudo-labels whose confidence scores are below a threshold and weight the importance of the remaining pseudo-labels equally, SIGN~\cite{sign} utilizes all pseudo annotations but assigns different loss weights according to the confidence scores of pseudo-labels.

\subsection{Zero-Shot Instance Segmentation (ZSIS)}
\label{subsec:zss-zsis}


Zheng \emph{et al.}~\cite{zsi} are the first to propose the task of zero-shot instance segmentation. They establish a simple mapping from visual features to semantic space then classify them using fixed semantic embeddings. The mapping is optimized by a mean-squared error reconstruction loss. Zheng \emph{et al.} also argue that disambiguation between background and unseen classes is crucial~\cite{sb-lab,blc}, they design a background-aware RPN and a synchronized background strategy to adaptively represent background.

%% file: sections/V4/5.ovd.tex
OVD removes the stringent restriction of inductive ZSD on the absence of unannotated novel objects as in~\cref{sec:zsd}. From this section on, we discuss methodologies resorting to weak supervision signals, \ie, the open-vocabulary setting in~\cref{sec:intro} and~\cref{sec:preliminaries}.


\subsection{Region-Aware Training}
\label{subsec:ovd-rat}

This methodology incorporates image-text pairs~\cite{cc,cc12m,coco-captions} into detection training phase. The vast vocabulary $\mathcal{C}_{T}$ in captions encompasses both $\mathcal{C}_B$ and $\mathcal{C}_N$, thus aligning proposals containing novel proposals with words containing novel classes improves classification on $\mathcal{C}_N$.

\noindent \textbf{Weakly-Supervised Grounding or Contrastive Loss.} This line of work establishes a coarse and soft correspondence between regions and words via the average of the following two symmetrical losses:
\begin{equation}
\label{eq:t2i}
    \mathcal{L}_{T\rightarrow I} = -\log \frac{\exp(sim(I,T))}{\sum_{I^{'} \in \mathcal{B}} \exp(sim(I^{'},T))},
\end{equation}
\begin{equation}
\label{eq:i2t}
    \mathcal{L}_{I\rightarrow T} = -\log \frac{\exp(sim(I,T))}{\sum_{T^{'} \in \mathcal{B}} \exp(sim(I,T^{'}))},
\end{equation}
where $\mathcal{B}$ represents the image-text batch. The similarity $sim(I, T)$ between image $I$ and caption $T$ is given by:
\begin{equation}
\label{eq:sim}
    sim(I,T)=\frac{1}{N_{T}}\sum_{i=1}^{N_{T}}\sum_{j=1}^{N_{I}}\alpha_{i,j}\langle e_{i}^T \cdot e_{j}^{I} \rangle,
\end{equation}
where $N_{T}$ and $N_{I}$ are the number of nouns in the caption and the number of proposals in the image, respectively. $e_{i}^{T}$ and $e_{j}^{I}$ are the text and region embedding typically encoded by the CLIP text encoder and detection head. The weight $\alpha_{i,j}$ is calculated by:
\begin{equation}
\label{eq:alpha}
    \alpha_{i,j}=\frac{\exp\langle e_{i}^T \cdot e_{j}^{I} \rangle}{\sum_{j^{'}=1}^{N_{I}}\langle e_{i}^T \cdot e_{j^{'}}^{I} \rangle}.
\end{equation}


By minimizing the distance of matched image-caption pairs, novel proposals and novel classes are aligned in a weakly-supervised manner (\emph{w/o} knowing the correspondence between proposals and words). OVR-CNN~\cite{ovr-cnn} first formulates the OVD task. Previous ZSD methods only train the vision-to-language projection layer from scratch on base classes, which is prone to overfitting. OVR-CNN learns the projection layer during pretraining on image-caption pairs using~\cref{eq:t2i} and~\cref{eq:i2t} to align image grids and words. Following OVR-CNN, LocOv~\cite{locov} introduces a consistency loss that regularizes image-caption similarities over a batch to be the same before and after the multi-modal fusion transformer. LocOv utilizes both region and grid features for measuring~\cref{eq:sim} while OVR-CNN only adopts the latter. However, the multi-modal masked language modeling~\cite{pixelbert,vilbert,vilt,albef} objective in OVR-CNN and LocOv tends to attend to similar global proposals covering many concepts for distinct masked words. To force the multi-modal fusion transformer focus more on exclusive proposals containing only one concept for different masked words, MMC-Det~\cite{mmc-det} drives the attention map over proposals be divergent for different masked words via the proposed divergence loss. DetCLIP~\cite{detclip} adds per-category definition from WordNet~\cite{wordnet} for CLIP text encoder to encode richer semantics. The category names and definitions are individually and parallelly encoded to avoid unnecessary interactions between category names. DetCLIPv2~\cite{detclipv2} selects a single region that best fits the current word via \emph{argmax} instead of aggregating all region features via softmax in~\cref{eq:alpha}. It excludes the $\mathcal{L}_{I \rightarrow T}$ in its bi-directional loss due to the partial labeling problem, \emph{i.e.}, the caption usually only describes salient objects in the image, hence most proposals can not find their matching words in the caption. WSOVOD~\cite{wsovod} recalibrates RoI features via input-conditional coefficients over dataset attribute prototypes to de-bias different distributions in different datasets. It employs multiple instance learning~\cite{mil} to address the lack of box annotations. 

Previous work perform weakly-supervised grounding only on relatively small-scale image-text pairs, another series of work pretrains the model on web-scale image-text datasets. RO-ViT~\cite{ro-vit} is pre-trained from scratch on the same dataset of ALIGN~\cite{align} using focal loss~\cite{focal} instead of cross-entropy loss to mine the hard negative examples. The positional embeddings are randomly cropped and resized to the whole-image resolution during pretraining, causing the model to regard pretrained images not as full images but as region crops from some unknown larger images. This matches the usage of proposals in the detection fine-tuning stage as they are cropped from a holistic image. To improve local feature representation for localization, CFM-ViT~\cite{cfm-vit} adds the masked autoencoder objective~\cite{mae} besides the bi-directional contrastive loss during pretraining on ALIGN~\cite{align} dataset. It randomly masks the positional embeddings during pretraining to achieve a similar effect of RO-ViT~\cite{ro-vit}. RO-ViT and CFM-ViT only pretrain the image encoder, while detection-specific components such as FPN~\cite{fpn} and RoI head~\cite{faster-rcnn} are randomly initialized and trained only on detection data. To bridge the architecture gap between pretraining and detection finetuning, DITO~\cite{dito} pretrains FPN~\cite{fpn} and RoI head~\cite{faster-rcnn} along with image encoder. Embeddings of randomly sampled regions are max pooled and image-text contrastive loss is applied on each FPN level separately.


\noindent \textbf{Bipartite Matching.} VLDet~\cite{vldet} formulates region-word alignment as a set-matching problem between regions and nouns. The matching is automatically learned on image-caption pairs via the off-the-shelf Hungarian algorithm~\cite{detr}. Following VLDet, GOAT~\cite{goat} mitigates the biased objectness score by comparing region features with an open corpus of external object concepts as another assessment of objectness. GOAT reconstructs the base classifier by taking a weighted mean of top-\emph{k} similar external concept embeddings \emph{w.r.t.} a base concept to enhance generalization. OV-DETR~\cite{ov-detr} conditions object queries on concept embeddings. They are constructed either from text embeddings or image embeddings by feeding base ground-truth boxes and novel proposals to the CLIP image encoder. It reformulates the set matching problem into conditional binary matching, which measures the matchability between detection outputs and the conditional object queries. However, the conditioned object queries are class-specific, \ie, the number is linearly proportional to the number of classes. Prompt-OVD~\cite{prompt-ovd} addresses this slow inference speed of OV-DETR by prepending class prompts instead of repeatedly adding to object queries and changing the binary matching objective to a multi-label classification cost. It further proposes RoI-based masked attention and RoI pruning to extract region embeddings in just one forward pass of CLIP. CORA~\cite{cora} learns region prompts following the addition variant of VPT~\cite{vpt} to adapt CLIP into region-level classification. The proposed anchor pre-matching makes object queries class-aware and can avoid repetitive per-class inference in OV-DETR~\cite{ov-detr}. Based on OV-DETR~\cite{ov-detr}, EdaDet~\cite{edadet} preserves fine-grained and generalizable local image semantics for attaining better base-to-novel generalization. 


\noindent \textbf{Leveraging Visual Grounding Datasets.} Only resorting to image-level captions brings noise and misalignment in region-word correspondence learning. This line of work leverages ground-truth region-level texts to help mitigate this problem. Each region-level description may contain one object (phrase grounding~\cite{flickr30k,vg}) or multiple objects with a subject (referring expression comprehension and segmentation~\cite{refcoco/+,refcocog}). These datasets are combined into one dataset termed GoldG in MDETR~\cite{mdetr}. It proposes soft token prediction to predict the span of tokens in these texts and performs contrastive loss at the region-word level in the latent feature space. Note that GLIP~\cite{glip} in~\cref{subsec:ovd-pl} also uses the GoldG dataset but its main purpose is to generate reliable pseudo labels for subsequent self-training. MAVL~\cite{mavl} improves MDETR by multi-scale deformable attention~\cite{deformable-detr} and late fusion between image and text modality. MQ-Det~\cite{mq-det} augments language queries in GLIP~\cite{glip} with fine-grained vision exemplars in a gated residual-like manner. It takes vision queries as keys and values to the class-specific cross-attention layer. The vision-conditioned masked language prediction forces the model to align with vision cues to reduce the learning inertia problem~\cite{mq-det}. YOLO-World~\cite{yolo-world} follows the formulation of GLIP and mainly focuses on equipping the YOLO series with open-vocabulary detection capability. SGDN~\cite{sgdn} also leverages grounding datasets~\cite{flickr30k,vg}, but it exploits additional object relations in a scene graph to facilitate discovering, classifying, and localizing novel objects.

\subsection{Pseudo-Labeling}
\label{subsec:ovd-pl}

Models advocating pseudo-labeling also leverage abundant image-text pairs as in~\cref{subsec:ovd-rat}, but additionally, they adopt large pretrained VLMs or themselves (via self-training) to generate pseudo labels. They need to know the exact novel categories $\mathcal{C}_N$ in the training stage. Detectors are then trained on the unification of base seen class annotations and pseudo labels. According to the type and granularity of pseudo labels, methods can be grouped into: pseudo region-word pairs, region-caption pairs, and pseudo captions. 


\noindent \textbf{Pseudo Region-Word Pairs.} The bi-directional grounding or contrastive loss in~\cref{subsec:ovd-rat} allows one region/word to correspond to multiple words/regions weighted by softmax. However, this line of work explicitly allows only one region/word to correspond to one word/region (\ie, hard alignment). RegionCLIP~\cite{regionclip} leverages CLIP to create pseudo region-word pairs to pretrain the image encoder of the detector. \emph{However, proposals with the highest CLIP scores yield low localization performance}. Targeting at this problem, VL-PLM~\cite{vl-plm} fuses CLIP scores with objectness scores and repeatedly applies the RoI head to remove redundant proposals. GLIP~\cite{glip} reformulates object detection into phrase grounding and trains the model on the union of detection and grounding data. It enables the teacher to utilize language context for grounding novel concepts, while previous pseudo-labeling methods only train the teacher on detection data. GLIPv2~\cite{glipv2} further reformulates visual question answering~\cite{vqa,bottom-up-top-down} and image captioning~\cite{show-attend-tell,scst} into a grounded VL understanding task. Following GLIP~\cite{glip}, Grounding DINO~\cite{grounding-dino} upgrades the detector into a Transformer-based one, enhancing the capacity of the teacher model. Instead of generating pseudo labels once, PromptDet~\cite{prompt-det} iteratively learns region prompts and sources uncurated web images in two rounds, leading to more accurate pseudo boxes in the second round. Similar to semi-supervised detection~\cite{ubt,soft-teacher}, SAS-Det~\cite{sas-det} also proposes to refine the quality of pseudo boxes online. The student is optimized on pseudo boxes and base ground-truths. Then, the teacher updates its parameters through the exponential moving average of the student. Thereby the quality of pseudo boxes is gradually improved during training. Previous approaches adopt a simple thresholding~\cite{glip,glipv2,grounding-dino} or a top-scoring heuristic rule~\cite{3ways,regionclip,rkdwtf,prompt-det} to filter pseudo labels, which is vulnerable and lacking explainability. PB-OVD~\cite{pb-ovd} employs GradCAM~\cite{grad-cam} to compute the activation map in the cross-attention layer of ALBEF~\cite{albef} \emph{w.r.t.} an object of interest in the caption. Then, all proposals that overlap the most with the activation map are regarded as pseudo ground-truths. CLIM~\cite{clim} combines multiple images into a mosaicked image with each image treated as a pseudo region, and the paired caption is deemed as the region label. CLIM can be applied to many open-vocabulary detectors~\cite{ovr-cnn,detic,baron}. To generate more accurate pseudo labels, VTP-OVD~\cite{vtp-ovd} introduces an adapting stage to enhance the alignment between pixels and categories via learnable visual~\cite{vpt} and textual prompts. ProxyDet~\cite{proxy-det} argues that, pseudo-labeling can not improve novel classes other than those defined in $\mathcal{C}_N$. It finds that many novel classes reside in the convex hull constructed by base classes in the CLIP visual-semantic embedding space. These novel classes can be approximated via a linear mixup between a pair of base classes. Hence, ProxyDet synthesizes region and text embeddings of proxy (fake) novel classes and aligns them to achieve a broader generalization ability beyond those pseudo-labeled novel classes. CoDet~\cite{codet} builds a concept group comprising all image-text pairs that mention the concept in their captions. The most common object appearing in the group is assumed to match the concept. Hence it reduces the problem of modeling cross-modality correspondence (region-word) to in-modality correspondence (region-region). The new formulation requires finding the most co-occurring objects by utilizing the fact that VLMs-IE exhibits feature consistency for visually similar regions across images.


\noindent \textbf{Pseudo Region-Caption Pairs.} Models in this category establish pseudo correspondence between the whole caption and a single image region, which is easier and less noisy compared to the more fine-grained region-word correspondence. Contrary to weakly-supervised detection that propagates image-level labels to corresponding proposals, Detic~\cite{detic} side-steps this error-prone label assignment process. It simply trains the max-size proposal to predict all image-level labels, similar to multiple instance learning~\cite{mil} and multi-label classification. The max-size proposal is assumed to be big enough to cover all image-level labels. Thus, the classifier encounters various novel classes during training on ImageNet21K~\cite{ilsvrc} and it can be generalized to novel objects at inference. Building on top of Detic, Kaul \emph{et al.}~\cite{mmc} employ a large language model, \emph{i.e.}, GPT-3~\cite{gpt-3}, to generate rich descriptions for text embeddings. Besides, vision exemplars are encoded and merged with text embeddings to incorporate in-modality classification clues. 3Ways~\cite{3ways} regards the top-scoring bounding box per image as correspondence to the whole caption. It also augments text embeddings to avoid overfitting and includes trainable gated shortcuts to stabilize training. PLAC~\cite{plac} learns a region-to-text mapping module that pulls close image regions with the corresponding caption. Then, novel proposals are matched to these region embeddings containing novel semantics.

\begin{figure}[t]
    \centering
    \includegraphics[height=4.43cm, width=8.9cm]{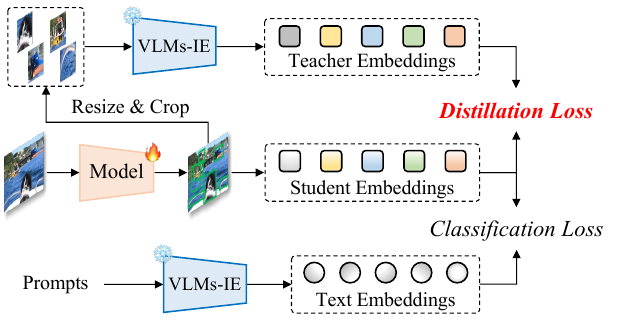}
    \vspace{-2em}
    \caption{A basic pipeline of knowledge distillation methodology. Distillation loss is typically a $\mathcal{L}_1$ loss. We omit the localization branch for brevity.}
    \label{fig:knowledge-distillation-framework}
\end{figure}


\noindent \textbf{Pseudo Captions.} PCL~\cite{pcl} proposes to generate another type of pseudo label, \emph{i.e.}, pseudo captions describing objects in natural language instead of generating bounding boxes. It leverages an image captioning model to generate captions for each object, which are then fed into the CLIP text encoder, encoding the class attributes and relationships with the surrounding environment. It can be regarded as better prompting compared to the template prompts in CLIP.

\subsection{Knowledge Distillation}
\label{subsec:ovd-kd}

Knowledge distillation methodology (\emph{c.f.}~\cref{fig:knowledge-distillation-framework}) employs a teacher-student paradigm where the student learns to mimic visual knowledge from the teacher image encoder, thus approaching the well-aligned visual-semantic space of the teacher more effectively. Current methods mainly vary on the granularity of the distillation region and type of distillation objective.


\noindent \textbf{Distillation at Single-Region Level.} Depending on the IoUs with base ground-truths and objectness scores, proposals can be categorized into base, novel, and background proposals. Two pioneering works ViLD~\cite{vild} and ZSD-YOLO~\cite{zsd-yolo} distill CLIP visual embeddings into Faster R-CNN and YOLOv5 detectors, respectively. ViLD first verifies that RPN well localizes novel proposals, regardless of trained either only on base annotations or the union of base and novel annotations. Hence, mimicking teacher visual embeddings of both base and novel proposals instead of only base ones improves the alignment between novel objects and novel classes. ZSD-YOLO also confirms the importance of applying distillation on these generalizable proposals instead of only on base ground truths. Though student visual embeddings resemble the teacher to an extent after distillation, there is still a gap between them. ViLD further ensembles the predictions of teacher and student via a weighted geometric mean during inference. The ensemble trusts teacher over student on novel predictions and vice versa on base predictions. Therefore it alleviates the overfitting of students to base classes, and it is inherited in many subsequent works. LP-OVOD~\cite{lp-ovod} employs a sigmoid focal loss~\cite{focal} (vs. softmax in ViLD) and constructs top relevant proposals of novel classes for training a novel classifier. In contrast to ViLD, EZSD~\cite{ezsd} argues that RPN is biased toward base classes. It regards predefined anchor boxes tiled over the image and then these boxes are filtered by CLIP score over base and novel classes as distillation regions. EZSD finetunes normalization layers of CLIP on detection data, then distills from this adapted CLIP. SIC-CADS~\cite{sic-cads} designs a multi-label recognition ranking loss for text alignment instead of the cross-entropy loss in previous work. The image-level scores are then used to weight instance-level scores to infer which objects may exist in the current image.


\noindent \textbf{Distillation at Bag-of-Regions Level.}
Distilling single region consisting of at most one object into student discards the knowledge of co-occurrence of visual concepts and their compositional structure implicitly captured in the CLIP image encoder. To harness this knowledge, BARON~\cite{baron} samples nearby regions to form multiple groups of bag-of-regions. These bag-of-regions embeddings are encoded by CLIP text encoder then contrasted with the image crops enclosing the bag-of-regions via InfoNCE loss~\cite{infonce}. 


\noindent \textbf{Distillation at Image-Region Level.} Beyond region level distillation, OADP~\cite{oadp} introduces a distillation pyramid that distills block-wise patches and a box covering the whole image. Compared to only using a single region of downstream images, this encourages the student to recover the full visual-semantic space of CLIP. GridCLIP~\cite{grid-clip} also aligns global image embeddings between two CLIP image encoders with one trainable and the other frozen.



\noindent \textbf{Distillation Objective Beyond $\mathcal{L}_1$ Loss.} Rasheed \emph{et al.}~\cite{rkdwtf} propose an inter-embedding relationship matching loss that forces student embeddings to share the same inter-embedding similarity structure as teacher embeddings. DK-DETR~\cite{dk-detr} argues that the $\mathcal{L}_1$ loss constrains each element of student and teacher visual embeddings to be the same which is too strict and poses training difficulty. It instead maximizes the similarity between student and teacher visual embeddings of the same object and pushes them far away for different objects via a binary cross-entropy loss. Since the caption consists of novel nouns, and the image consists of novel objects, HierKD~\cite{hierkd} contrasts the caption representations from the teacher with the ones from the student to improve base-to-new generalization. Another representative work DetPro~\cite{detpro} focuses on learning continuous context prompts~\cite{coop} following ViLD via a cross-entropy loss. Concretely, DetPro forces all novel and background proposals equally unlike any base class, thus learned context prompts avoid classifying novel proposals into base classes. In addition, DetPro divides foreground proposals into disjoint groups according to their IoUs \emph{w.r.t.} ground-truths. Prompt representations for each group are learned separately to describe different levels of contexts accurately, \emph{e.g.}, the learned prompts may be ``A partial photo of [CLASS].'' or ``A complete photo of [CLASS]''. CLIPSelf~\cite{clipself} discovers that, in terms of classification accuracy, for ViT-based CLIP image encoders, using RoIAlign~\cite{mask-rcnn} to crop the ground-truths directly on the dense feature maps performs much inferior than forwarding the image crops to CLIP image encoder. In fact, F-VLM~\cite{f-vlm} only applied to ResNet-based CLIP image encoders. CLIPSelf rectifies this issue for ViT-based CLIP via a cosine self-distillation loss between the holistic image crop representation and patch-level dense presentations. 


\subsection{Transfer Learning}
\label{subsec:ovd-tl}


Knowledge distillation separates VLMs-IE and detector backbone during training, while transfer learning-based models discard the detector backbone and only utilize VLMs-IE. For example, fine-tuning VLMs-IE on detection data, or extracting visual features via the frozen VLMs-IE. OWL-ViT~\cite{owl-vit} removes the final token pooling layer of the CLIP image encoder and attaches a lightweight detection head to each Transformer output token. Then it fine-tunes the whole model on detection data through a bunch of dedicated finetuning techniques in an end-to-end manner. UniDetector~\cite{unidetector} also trains the whole model, the image backbone is initialized with RegionCLIP~\cite{regionclip}. During training, UniDetector leverages images of multiple sources and heterogeneous label spaces. During inference, it calibrates the base and novel predictions via a class-specific prior probability recording how the network biases towards that category. Instead of fine-tuning the whole model, F-VLM~\cite{f-vlm} leverages the frozen CLIP image encoder to extract features and only trains the detection head. It ensembles predictions of the detector and CLIP via the same geometric mean (dual-path inference) as ViLD~\cite{vild}. Another line of work only employs the CLIP text encoder for open-vocabulary classification and discards the CLIP image encoder. ScaleDet~\cite{scaledet} unifies the multi-dataset label spaces by relating labels with sematic similarities across datasets. OpenSeeD~\cite{openseed} unifies OVD and OVS in one network. It proposes decoupled foreground-background decoding and conditioned mask decoding to compensate for task and data discrepancies, respectively. CRR~\cite{drr} analyzes whether the classification in RoI head network hampers the generalization ability of RPN and proposes to decouple them, \emph{i.e.}, RPN and RoI head do not share the backbone and they are separately trained. Sambor~\cite{sambor} introduces a ladder side adapter which assimilates localization and semantic prior from SAM and CLIP simultaneously. The automatic mask generation in SAM~\cite{sam} is employed to enhance the robustness of class-agnostic proposal generation in RPN.

%% file: sections/V4/6.ovs.tex
In this section, we review semantic, instance, and panoptic segmentation tasks using the same taxonomy in~\cref{sec:ovd}.


\subsection{Open-Vocabulary Semantic Segmentation (OVSS)}
\label{subsec:ovs-ovss}

\subsubsection{Region-Aware Training}
\label{subsubsec:ovs-ovss-rat}


\noindent \textbf{Weakly-Supervised Grounding or Contrastive Loss.} Using the same~\cref{eq:i2t} and~\cref{eq:t2i}, OpenSeg~\cite{openseg} randomly drops each word to prevent overfitting. SLIC~\cite{slic} incorporates local-to-global self-supervised learning in DINO~\cite{dino,dinov2} to improve the quality of local features for dense prediction.


\noindent \textbf{Learning from Natural Language Supervision Only.} Methods fall into this category aim to learn transferrable segmentation models purely on image-text pairs without densely annotated masks. GroupViT~\cite{group-vit} progressively groups segment tokens into arbitrary-shaped and semantically-coherent segments given their assigned group index via gumbel-softmax~\cite{gumbel-softmax}. Segment tokens in the last layer are average pooled and contrasted with captions like CLIP~\cite{clip}. Besides the image-text contrastive loss, ViL-Seg~\cite{vil-seg} incorporates the same local-to-global correspondence learning in DINO~\cite{dino} via a multi-crop strategy~\cite{multi-crop} to capture fine-grained semantics. An online clustering head trained by mutual information maximization~\cite{mi} groups pixel embeddings at the end of ViT. Following GroupViT, SegCLIP~\cite{segclip} enhances local feature representation with additional MAE~\cite{mae} objective and a superpixel-based Kullback-Leibler loss. OVSegmentor~\cite{ovsegmentor} adopts slot attention~\cite{slot-attn} to bind patch tokens into groups. To ensure visual invariance across images for the same object, OVSegmentor proposes a cross-image mask consistency loss. CLIP ViT-based image encoder performs worse on patch-level classification. To remedy this issue, PACL~\cite{pacl} adds an additional patch-text contrastive loss between patch embeddings and the caption. TCL~\cite{tcl} designs a region-level text grounder that produces text-grounded masks containing objects only described in the caption without irrelevant areas. Then the matching is performed on the grounded image region and text instead of the whole image and text. SimSeg~\cite{simseg} points out that CLIP heavily relies on contextual pixels and contextual words instead of entity words. Hence, instead of densely aligning all image patches with all words, SimSeg sparsely samples a portion of patches and words used for the bi-directional contrastive loss in~\cref{eq:i2t} and~\cref{eq:t2i}. 

\subsubsection{Pseudo-Labeling}
\label{subsubsec:ovs-ovss-pl}


Zabari \emph{et al.}~\cite{ttd} leverage an interpretability method~\cite{transformer-interpretability} to generate a coarse relevance map for each category. The relevance map is refined by test-time-augmentation techniques (\eg, horizontal flip, contrast change, and crops). The synthetic supervision is generated from the refined relevance maps using stochastic pixel sampling. 

\subsubsection{Knowledge Distillation}
\label{subsubsec:ovs-ovss-kd}


GKC~\cite{gkc} enriches the template prompts with synonyms from WordNet~\cite{wordnet} instead of relying only on a single category name to guess what the object looks like. The text-guided knowledge distillation transfers the inter-class distance relationships in semantic space into visual space, which is similar to~\cite{rkdwtf}. SAM-CLIP~\cite{sam-clip} merges the image encoders of SAM~\cite{sam} and CLIP~\cite{clip} into one via cosine distillation loss in a memory replay and rehearsal way on a subset of their pretraining images. It combines the superior localization ability of SAM and semantic understanding ability of CLIP. ZeroSeg~\cite{zeroseg} builds on top of MAE~\cite{mae} and GroupViT~\cite{group-vit}. Unlike GroupViT which requires text supervision, ZeroSeg distills the multi-scale CLIP image features into learnable segment tokens purely on unlabled images. The image is divided into multiple views thus capturing both local and global semantics.

\begin{figure}[t!]
    \centering
    \includegraphics[width=8.9cm, height=9.39cm]{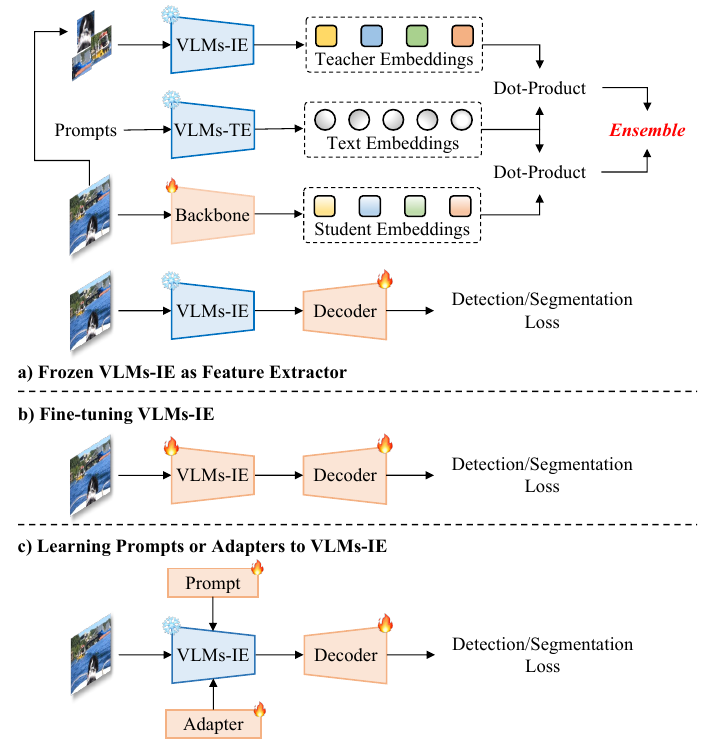}
\vspace{-2em}
    \caption{Framework for transfer learning-based models.}
    \label{fig:transfer-learning-based}
\vspace{-1em}
\end{figure}

\subsubsection{Transfer Learning}
\label{subsubsec:ovs-ovss-tl}

This methodology aims to transfer the VLMs-TE and VLMs-IE to segmentation tasks. The transfer strategy is explored in the following aspects: 1) only adopting VLMs-TE for open-vocabulary classification; 2) leveraging frozen VLMs-IE as a feature extractor; 3) directly fine-tuning VLMs-IE on segmentation datasets; 4) employing visual prompts or attaching a lightweight adapter to frozen VLMs-IE for feature adaptation. A detailed comparison is given in~\cref{fig:transfer-learning-based}.


\noindent \textbf{VLMs-TE as Classifier.} LSeg~\cite{lseg} simply replaces the learnable classifier in segmentor~\cite{dpt} with text embeddings from the CLIP text encoder. SAZS~\cite{sazs} focuses on improving boundary segmentation supervised by the output of edge detection on ground-truth masks. During inference, SAZS fuses the predictions with eigen segments obtained through spectral analysis on DINO~\cite{dino} to promote shape-awareness. Son \emph{et al.}~\cite{cel} align object queries and text embeddings by bipartite matching~\cite{detr}. They force queries not matched to any base or novel class to predict a uniform distribution over base and novel classes, thus avoiding the use of ``background'' embedding. A multi-label ranking loss is employed to encourage the similarity of any positive label to be higher than that of any negative label.


\noindent \textbf{Frozen VLMs-IE as Feature Extractor.} As shown in~\cref{fig:transfer-learning-based}, there are two variants of leveraging frozen VLMs-IE as feature extractor. The first is adding a parallel branch to the detector backbone, the predictions of two branches are later ensembled using geometric mean~\cite{vild,f-vlm}. The second discards the trainable detector backbone and utilize the frozen VLMs-IE as the only feature extractor. 

For the first variant, ZegFormer~\cite{zegformer} forwards masked image crops to frozen CLIP image encoder and ensembles its predicted scores via geometric mean~\cite{vild,f-vlm} with CLIP during inference. ReCo~\cite{reco} first retrieves an archive of exemplar images for each class from unlabelled images. Then, it leverages MoCo~\cite{moco} to extract seed pixels across exemplar images to construct reference image embeddings as $\text{1} \times \text{1}$ convolution classifier. The prediction is ensembled with DenseCLIP~\cite{denseclip}. SCAN~\cite{scan} injects semantic prior of CLIP into proposal embeddings to avoid biasing toward base classes. To mitigate the domain shift between natural images and masked crops, SCAN replaces background token embeddings in the feature map with CLIP [CLS] token.

For the second variant, ZSSeg~\cite{zsseg} adopts the same architecture as ZegFormer except that ZSSeg adopts learnable prompts similar to CoOp~\cite{coop} without ensembling with CLIP. MaskCLIP~\cite{maskclip} removes the query and key projection in the final attention pooling layer of CLIP image encoder to enhance local semantic consistency and suppress noisy context information. This trick has been utilized in many subsequent works. CLIP-DINOiser~\cite{clip-denoiser} performs a concept-aware linear combination of patch features guided by DINO~\cite{dino} to suppress the noisy features~\cite{maskclip,clip-surgery} in CLIP. It also leverages an unsupervised segmentation method~\cite{found} to produce a background map that rectifies false background assignment. MVP-SEG~\cite{mvp-seg} proposes multi-view prompt learning optimized by orthogonal contrastive loss to focus on different object parts. Peekaboo~\cite{peekaboo} explores how off-the-shelf Stable Diffusion (SD)~\cite{stable-diffusion} can perform grouping pixels with the proposed dream loss. OVDiff~\cite{ovdiff} is also entirely training-free, the same as Peekaboo~\cite{peekaboo}, it relies only on SD~\cite{stable-diffusion}. OVDiff removes the cross-modality similarity measurement, it directly compares against image features with part-, instance-, and class-level prototypes (support set) sampled from SD. FOSSIL~\cite{fossil} also transforms cross-modal comparison into uni-modal comparison by constructing visual prototypes from DINOv2~\cite{dinov2}. POMP~\cite{pomp} condenses semantic concepts over twenty-thousand classes into the learned prompts. It introduces local contrast and local correction strategy to reduce the memory consumption of CoOp~\cite{coop}. AttrSeg~\cite{attrseg} decomposes each category name into attributes, \emph{e.g.}, color, shape, texture, and parts, then hierarchically aggregates attribute tokens into the class embedding. In this way, the lexical ambiguity, neologism, and unnameability caused by the limited expressivity of a single class name are solved by providing any number of attributes describing the object. PnP-OVSS~\cite{pnp-ovss} pursues a training-free paradigm where GradCAM~\cite{grad-cam} plus cross-attention saliency map are combined to locate the object. An iterative saliency dropout is proposed to trade-off between over-segment (patches unrelated to the class are segmented in cross-attention map) and under-segment (GradCAM often narrowly focuses on the most discriminative patches while ignoring other useful patches). TagAlign~\cite{tagalign} introduces a multi-tag classification loss besides the image-text contrastive loss. Self-Seg~\cite{self-seg} leverages the feature consistency of BLIP~\cite{blip,blip2} to cluster image patches into distinct clusters then uses the captioner to generate open-ended descriptions. These captions are parsed into nouns and sent to X-Decoder~\cite{x-decoder} for segmentation. The nouns in generated captions may not exist in downstream mask labels. To resolve this issue, Self-Seg employs LLaMA2~\cite{llama2} to map non-existent parsed nouns into the most relevant pre-defined categories in the dataset.


\noindent \textbf{Fine-tuning VLMs-IE.} This group finetunes the CLIP image encoder shown in~\cref{fig:transfer-learning-based} to adapt its feature representations to segmentation tasks. DenseCLIP~\cite{denseclip} learns pixel-text matching directly on the feature map using densely annotated masks. OVSeg~\cite{ovseg} fine-tunes CLIP image encoder on the constructed mask-category pairs to address the domain gap between masked image crops with blank areas and natural images used to pretrain CLIP. CAT-Seg~\cite{cat-seg} devises a cost aggregation module including spatial and class aggregation to produce the segmentation map. It only fine-tunes the attention layers as finetuning all parameters of CLIP harms its open-vocabulary capabilities. Since ViT-based CLIP image encoders exihibit weak local region classification ability~\cite{f-vlm,clipself}, SED~\cite{sed} also utilizes the cost map as CAT-Seg with a ConvNext~\cite{convnext} backbone. MAFT~\cite{maft} fine-tunes CLIP image encoder by mimicking the IoU between proposals and ground-truths to make classfication mask-aware.


\noindent \textbf{Learning Prompts or Adapters for VLMs-IE.} Compared to fine-tuning CLIP image encoder, learning prompts or adapters can better preserve the generalization ability in novel classes. ZegCLIP~\cite{zegclip} and TagCLIP~\cite{tagclip} both adopt deep prompt tuning~\cite{vpt}. Additionally, TagCLIP designs a learnable trusty token generating trusty maps used to measure the reliability of the raw segmentation map by CLIP. CLIPSeg~\cite{clipseg} attaches a lightweight decoder to CLIP image encoder with U-Net~\cite{u-net} skip connections conditioned on text embeddings. SAN~\cite{san} attaches a lightweight vision transformer to the frozen CLIP image encoder. It requires only a single forward pass of CLIP. SAN decouples the mask proposal and classification stage by predicting attention biases applied to deeper layers of CLIP for recognition. CLIP Surgery~\cite{clip-surgery} discovers that CLIP has opposite visualizations similar to the findings of SimSeg~\cite{simseg} and noisy activations. The proposed architecture surgery replaces the query-key self-attention with a value-value self-attention to causes the opposite visualization problem. The feature surgery identifies and removes redundant features to reduce noisy activations. Instead of learning visual prompts, CaR~\cite{car} puts a red circle on the proposal as a prompt to guide the attention of CLIP toward the region~\cite{red-circle}. It designs a training-free and RNN-like framework where text queries are deemed as hidden states and are gradually refined to remove categories not present in the image. However, repeatedly forwarding the whole image multiple times incurs a slow inference speed.

\subsection{Open-Vocabulary Instance Segmentation (OVIS)}
\label{subsec:ovs-ovis}


\textbf{Region-Aware Training.} CGG~\cite{cgg} achieves the region-text alignment via a grounding loss, but not with the whole caption as in OVR-CNN~\cite{ovr-cnn}. CGG extracts object nouns so that object-unrelated words do not interfere with the matching process. In addition, CGG proposes caption generation to reproduce the caption paired with the image. $\text{D}^2\text{Zero}$~\cite{d2zero} proposes an unseen-constrained feature extractor and an input-conditional classifier to address the bias issue. It proposes image-adaptive background representations to discriminate novel and background classes more effectively.


\noindent \textbf{Pseudo-Labeling.} XPM~\cite{xpm} first trains a teacher model on base annotations, then self-trains a student model. The pseudo regions are selected as the most compatible region \emph{w.r.t} object nouns in the caption. However, pseudo masks contain noises that degrade performance. XPM assumes that each pixel in pseudo masks is corrupted by a Gaussian noise, and the student is trained to predict the noise level to down weight incorrect teacher predictions. Mask-free OVIS~\cite{mask-free-ovis} performs iterative masking using ALBEF~\cite{albef} and GradCAM~\cite{grad-cam} to generate pseudo-instances both for base and novel categories. To alleviate the overfitting issue, it avoids training base categories using strong supervision and novel categories using weak supervision. MosaicFusion~\cite{mosaicfusion} runs the T2I diffusion model on a mosaic image canvas to generate multiple pseudo instances simultaneously. Pseudo masks are obtained by aggregating cross-attention maps across heads, layers, and time steps.


\noindent \textbf{Knowledge Distillation.} OV-SAM~\cite{ov-sam} proposes to combine CLIP and SAM into one unified architecture. Adapters are inserted at the end of the CLIP image encoder and mimic SAM features via a mean-squared error loss. CLIP features are also fed into the SAM mask decoder to enhance open-vocabulary recognition ability.


\subsection{Open-Vocabulary Panoptic Segmentation (OVPS)}
\label{subsec:ovs-ovps}


\textbf{Region-Aware Training.} Uni-OVSeg~\cite{uni-ovseg} employs bipartite matching on image-caption pairs similar to VLDet~\cite{vldet} except that Uni-OVSeg designs a multi-scale cost matrix to achieve more accurate matching. X-Decoder~\cite{x-decoder} proposes an image-text decoupled framework for not only open-vocabulary panoptic segmentation but also other vision-language tasks. It defines latent and text queries responsible for pixel-level segmentation and semantic-level classification, respectively. APE~\cite{ape} jointly trains on multiple detection and grounding datasets as well as SA-1B~\cite{sam}. In contrast to GLIP~\cite{glip}, it reformulates visual grounding as object detection and independently encodes concepts instead of concatenating concepts and encoding them all. It treats stuff class as multiple disconnected standalone regions, removing separate head networks for things and stuff as in OpenSeeD~\cite{openseed}. Moreover, SA-1B is semantic-unaware, \ie, thing and stuff are indistinguishable. APE can utilize SA-1B for training with only one head without differentiating things and stuff.


\noindent \textbf{Knowledge Distillation.} Previous synthesizers~\cite{gan,wgan,cvae} in~\cref{subsubsec:zss-zsss-nvfs} with several linear layers do not consider the feature granularity gap between image and text modality. PADing~\cite{pading} proposes learnable primitives to reflect the rich and fine-grained attributes of visual features, which are then synthesized via weighted assemblies from these abundant primitives. In addition, PADing~\cite{pading} decouples visual features into semantic-related and semantic-unrelated parts and it only aligns the semantic-related parts to the inter-class relationship structure in the semantic space. 


\noindent \textbf{Transfer Learning.} FC-CLIP~\cite{fc-clip} resembles closely with F-VLM~\cite{f-vlm} in that both use a frozen CNN-based CLIP image encoder and take a geometric ensemble for base and novel classes separately. FreeSeg~\cite{freeseg} learns prompts separately for semantic/instance/panoptic tasks and different classes during the training stage. In inference, FreeSeg optimizes class prompts following test-time adaption~\cite{tta,ttp} by minimizing the entropy. PosSAM~\cite{possam} employs the frozen CLIP and SAM image encoder and fuses their output visual features via cross-attention. MasQCLIP~\cite{masqclip} follows MaskCLIP~\cite{maskclip} except that the query projection of the mask class token is learnable. This is to reduce the shift of CLIP from extracting image-level features to classifying a masked region in an image. OMG-Seg~\cite{omg-seg} explores the paradigm of unifying semantic/instance/panoptic segmentation and their counterparts in videos under both closed and open vocabulary settings. Semantic-SAM~\cite{semantic-sam} consolidates multiple datasets across granularities and trains on decoupled object and part classification, achieving semantic-awareness and granularity-abundance through a multiple-choice learning paradigm. ODISE~\cite{odise} resorts to T2I diffusion models~\cite{stable-diffusion} as the mask feature extractor. It also proposes an implicit captioner via the CLIP image encoder to map images into pseudo words. The training is driven by a bi-directional grounding loss in~\cref{subsec:ovd-rat}. Same as ODISE~\cite{odise}, HIPIE~\cite{hipie} ensembles classification logits with CLIP. It can hierarchically segment things, stuff, and object parts. It employs two separate decoders for things and stuff to deal with different losses. Zheng~\emph{et al.}~\cite{maskclip2} design mask class tokens to extract dense image features corresponding to each mask area via the proposed relative mask attention. OPSNet~\cite{opsnet} attaches spatial adapter on top of CLIP image encoder. Since mask embeddings are reliable at predicting base classes, and CLIP image embeddings preserve novel class recognition ability, thus OPSNet modulates the two to learn from each other.

%% file: sections/V4/7.ov3d.tex
\subsection{Open-Vocabulary 3D Scene Understanding}
\label{subsec:ov3d-pc}


Open-vocabulary 3D scene understanding is relatively under-explored and suffers a more severe data scarcity issue, even pairing point clouds with text descriptions is not feasible. Hence, the point-cloud and text modality are typically bridged via the intermediate image modality, where VLMs (\emph{e.g.}, CLIP) step in to guide the association. Typically, the dataset annotates the projection matrix that transforms 3D point clouds into 2D boxes and vice versa.

\subsubsection{Open-Vocabulary 3D Object Detection (OV3D)}
\label{subsubsec:ov3d-od}


OV-3DET~\cite{ov-3det} leverages pseudo 2D boxes \emph{w/o} class labels generated from a pretrained 2D open-vocabulary detector~\cite{detic}. These pseudo 2D boxes are then back projected to 3D space, which is deemed as the supervision signal for localization learning. To classify the predicted 3D RoI features, they are first back projected to image crops which are then encoded by CLIP image encoder. These image crop features are assigned a semantic label by CLIP, thus the 3D RoI feature can connect to its labeled text embedding via paired relationship between 3D and 2D space. A triplet contrastive loss is applied to drive the 3D RoI feature to approach both the projected image feature and its associated text embedding. FM-OV3D~\cite{fm-ov3d} follows a similar pipeline by back projecting 2D boxes of Grounded-SAM~\cite{grounded-sam}. For open-vocabulary 3D recognition, FM-OV3D aligns 3D RoI features with CLIP representations of both text and visual prompts from GPT-3~\cite{gpt-3} and Stable Diffusion~\cite{stable-diffusion}. OpenSight~\cite{opensight} increases temporal awareness that correlates the predicted 3D boxes across consecutive timestamps and recalibrates missed or inaccurate boxes. In contrast to relying on a 2D open-vocabulary detector, CoDA~\cite{coda} discovers novel 3D objects in an online and progressive manner, even though the detector is trained only on a few 3D base annotations. The projected 2D box features are used to distill CLIP knowledge into 3D object features. Though the image modality bridges the intermediate representation between point clouds and the text modality, it requires extra alignment between point clouds and images which may limit the performance. L3Det~\cite{object2scene} proposes not to leverage images but large-scale large-vocabulary 3D object datasets. It inserts 3D objects covering both base and novel classes into the scene in a physically reasonable way and generates grounded descriptions for them. Therefore, L3Det bypasses the image modality and directly learns the alignment between 3D objects and texts through contrastive learning.

\subsubsection{Open-Vocabulary 3D Segmentation}
\label{subsubsec:ov3d-seg}


\noindent \textbf{Open-Vocabulary 3D Semantic Segmentation (OV3SS).} SeCondPoint~\cite{secondpoint} and 3DGenZ~\cite{3dgenz} basically follow the pipeline of novel visual feature synthesis in~\cref{subsec:zsd-nvfs}. OpenScene~\cite{openscene} embeds point features into CLIP latent space, minimizing the differences with the aggregated pixel features via a distillation loss. Thus, by aligning point features with pixel features which in turn aligned with text embeddings, point features are aligned with text embeddings. PLA~\cite{pla} first calibrates predictions to avoid over-confident scores on base classes. Then, hierarchical coarse-to-fine point-caption pairs, \emph{i.e.}, scene-, view-, and entity-level point-caption association via a pretrained captioning model~\cite{gpt-2} are constructed, effectively facilitating learning from language supervisions. However, the pseudo-captions at the view level only cover sparse and salient objects in a scene, failing to provide fine-grained descriptions. To enable dense regional point-language associations, RegionPLC~\cite{region-plc} captions image patches in a sliding window fashion together with object proposals. A point-discriminative contrastive learning objective is further proposed that makes the gradient of each point unique. Note that both PLA and RegionPLC are capable of instance segmentation.


\noindent \textbf{Open-Vocabulary 3D Instance Segmentation (OV3IS).} OpenMask3D~\cite{openmask3d} aggregates per-mask feature via multi-view fusion of CLIP image embeddings. The projected 2D segmentation masks are refined by SAM~\cite{sam} to remove outliers. MaskClustering~\cite{mask-clustering} proposes multi-view consensus rate to assess whether 2D masks from off-the-shelf detectors belong to the same 3D instance (\ie, they should be merged). An iterative graph clustering is designed to better distinguish distinct 3D instances in a class-agnostic manner. OpenIns3D~\cite{openins3d} requires no image inputs as a bridge between point clouds and texts. It synthesizes scene-level images and leverages OVD detector in 2D domain to detect objects the associates them with semantic labels. Open3DIS~\cite{open3dis} addresses the problem of proposing high-quality small-scale and geometrically ambiguous instances by aggregating them across frames.

\subsection{Open-Vocabulary Video Understanding (OVVU)}
\label{subsec:ov3d-vis}


OV2Seg~\cite{ov2seg} first proposes open-vocabulary video instance segmentation task that simultaneously detects, segments, and tracks arbitrary instances regardless of their presence in the training set. It collects a large vocabulary video instance segmentation dataset covering 1,212 categories for benchmarking. OV2Seg leverages CLIP text encoder to classify queries from its memory-induced tracking module. A concurrent work OpenVIS~\cite{openvis} first proposes instances in a frame exhaustively. Then in the second stage, it designs square crop that avoids distorting the aspect ratio of instances to better conform to the pre-processing of the CLIP image encoder. However, previous work~\cite{ov2seg,openvis} align the same instance in different frames separately with text embeddings without considering the correlation across frames. BriVIS~\cite{brivis} links instance features across frames as a Brownian bridge and aligns the bridge center with text embeddings. DVIS++~\cite{dvis++} presents a unified framework capable of various video segmentation simultaneously. 

%% file: sections/V4/8.challenges-outlook.tex
\subsection{Challenges}
\label{subsec:challenges}

\textbf{Overfitting on Base Classes.} Due to the lack of novel annotations, the overfitting or bias issue is severe and manifested in three folds: 1) Base proposals are of higher quality and quantity than novel proposals. 2) Novel proposals are prone to be misclassified into base classes. 3) Recognition confidence on base classes is much higher than novel classes. For the first aspect, many endeavors seek to 1) adopt a standalone and frozen proposal generator to avoid classification of base classes in detector head affecting gradients of proposal generator~\cite{drr}; 2) employ pure localization quality-based objectness score~\cite{oln} \emph{w/o} foreground-background binary classification, or design complementary objectness measures utilizing a large corpus of concepts~\cite{goat}. Leveraging unsupervised localization methods~\cite{uol-survey} built on DINO~\cite{dino} or SAM~\cite{sam} for proposal generation can also potentially mitigate the bias. For the second aspect, recalibration in the inference stage is used in many works~\cite{vild,f-vlm} by separating and ensembling the predictions of base and novel classes between the detector and CLIP. 

\noindent \textbf{Confusion on Novel and Background Concepts}. Since only base and background text embeddings are used as classifier weights during training, both novel and background proposals are classified as background. This drawback will cause novel proposals misclassified into the background in inference. Besides, background text embedding is typically encoded by passing the template prompt ``A photo of the [background].'' into a CLIP text encoder or an all-zero vector. This simple representation is not sufficient and representative to cope with diverse contexts.

\noindent \textbf{Correct Region-Word Correspondence.} Though image-text pairs are cheap and abundant, the region-word correspondence is weak, noisy, and explicitly unknown. The bi-directional grounding loss in~\cref{subsec:ovd-rat} may cheat on establishing correct region-word correspondence by only aligning bag-of-regions to bag-of-words. Besides, the object nouns in the caption may only cover salient objects, and they are far less than the number of proposals, \ie, many objects may not find the matching words. Pseudo labels impose the constraint of one region connecting to one word and vice versa. However, they are generated once and done, iteratively refining the quality of pseudo labels in online training is less explored.

\noindent \textbf{Large VLMs Adaptation.} \emph{There is a distinct discrepancy and domain gap in terms of image resolution, context, and task statistics between the pretraining and detection tuning phases}. During the pertaining phase, CLIP receives low-resolution images with full contexts including object occurrences, relationships, spatial layout, \emph{etc}. However, in the detection tuning phase, CLIP either receives high-resolution images or low-resolution masked image crops containing a single object without any context. The masked image crops are of non-square sizes or extreme aspect ratios, the pre-processing step of CLIP resizes the shorter edge and center crop, adds more distortion and aggravates this gap. The prediction of CLIP is also not sensitive to localization quality, \emph{i.e.}, given an image crop with only a small portion of objects of interest, CLIP still makes predictions with high confidence. Besides, fully finetuning the whole VLMs for adaptation always leads to catastrophic forgetting of prior knowledge on open-vocabulary tasks. In light of this, lightweight adapters or prompt tuning plays a crucial role in large VLMs adaptation.

\noindent \textbf{Inference Speed and Evaluation Metrics.} Current OVD and OVS methods mainly build on top of mainstream object detectors, such as Faster R-CNN~\cite{faster-rcnn}, DETR~\cite{detr}, and Mask2Former~\cite{mask2former}, which are slow when deployed on edge devices. However, lightweight detectors like YOLO~\cite{yolov3} may aggravate the above challenges. Real-time detectors result in lower recall rate of novel objects. Besides, distilling the knowledge of large VLMs into these small-scale models remains questionable due to their limited learning capacity. Meanwhile, the evaluation is also problematic~\cite{ovs-metric}, for example, suppose the predicted category and label are synonyms, current metrics will not deem the prediction as a true positive. This might be too strict given the fact that in an open-world, many words are interchangeable.

\subsection{Future Directions}
\label{subsec:outlook}

\noindent \textbf{Enabling Open-Vocabulary on Other Scene Understanding Tasks.} Currently, other tasks including open-vocabulary 3D scene understanding, video analysis~\cite{gao2023compositional}, action recognition, object tracking, and human-object interaction~\cite{li2023zero}, \emph{etc}, are underexplored. In these problems, either the weak supervision signals are absent or the large VLMs yield pool open-vocabulary classification ability. Enabling open-vocabulary beyond detection and segmentation has became a mainstream trend.

\noindent \textbf{Unifying OVD and OVS.} Unification is an inevitable trend for computer vision. Though there are several works addressing different segmentation tasks simultaneously~\cite{freeseg,dataseg,odise,hipie} or training on multiple detection datasets~\cite{unidetector,scaledet}. A universal foundational model for all tasks and datasets~\cite{openseed} remains barely untouched, or even further, accomplishing 2D and 3D open-vocabulary perception simultaneously can be more challenging.

\noindent \textbf{Multimodal Large Language Models (LLMs) for Perception.} Multimodal large language models~\cite{contextdet,detgpt,visionllm,lisa} typically comprise three parts: 1) a vision encoder; 2) a mapper that maps the visual features to the input space of LLM.2) an LLM for decoding desired outputs. Bounding boxes are represented as two corner integer points~\cite{pix2seq} and similarly for segmentation masks via sampling points on the contour~\cite{seqtr,pix2seq2} of the mask. The reasoning capability of user intentions and interactive detection within a language context endow multi-modal LLMs for detection and segmentation in the wild.

\noindent \textbf{Combining Large Foundation Models.} Different foundation models have different capabilities. SAM~\cite{sam} excels in localizing objects but in a class-agnostic manner. CLIP~\cite{clip} is superior at image-text alignment but behaves like bag-of-words and lacks spatial-awareness for dense prediction tasks. DINO~\cite{dino,dinov2} exhibits a superior cross-image correspondence for objects or parts of the same class but is mainly used in unsupervised localization tasks. T2I Diffusion models generate astonishing images but their usage in discriminative dense prediction tasks remains under-explored. In a nutshell, how to benefit from these emerging large foundation models and combine them are key questions for future research.

\noindent \textbf{Real-Time OVD and OVS.} Current models possess a heavy backbone and neck architecture, which are unsuitable for real-time applications. To fully unleash the productive potential of OVD and OVS, exploring real-time detectors~\cite{yolo-world,rt-detr} with open-vocabulary recognition ability is a promising research direction.

%% file: sections/V4/9.conclusion.tex
We covered a broad and concrete development of OVD and OVS in this survey. First, the background section consistig of definitions, related domains and tasks, canonical closed-set detectors and segmentors, and large VLMs were introduced. Then, we detailed near two hundred OVD and OVS methods. At the task level, both 2D detection and different segmentation tasks are discussed, along with 3D scene and video understanding. At the methodology level, we pivoted on the permission and usage of weak supervision signals and grouped most of the existing methods into six categories, which are universal across tasks. In the end, challenges and promising directions are discussed to facilitate future research. In addition, we benchmarked the performance of state-of-the-art methods along with their vital components for each task in the appendix.

%% file: sections/V4/bio.tex
\vspace{-40em}

\begin{IEEEbiography}
[{\includegraphics[width=1in,height=1.25in,clip,keepaspectratio]{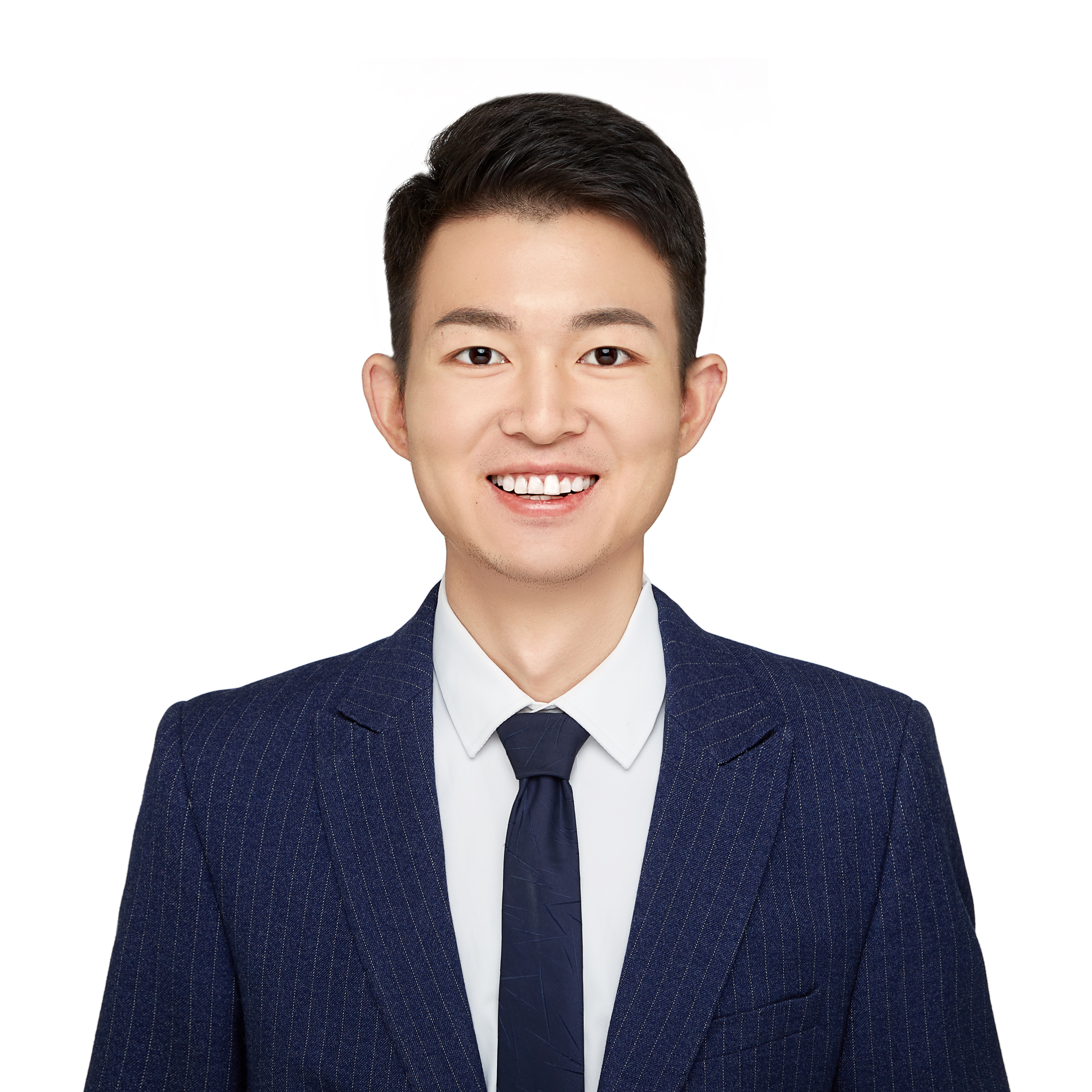}}]{Chaoyang Zhu} currently is a Ph.D. student at the Department of Computer Science and Engineering, HKUST. He received the M.Sc degree in Computer Technology from Xiamen University in 2023, and the B.Eng. degree in Computer Science and Technology from Hangzhou Dianzi University in 2019. His research interests are computer vision and multimodal learning. 
\end{IEEEbiography}

\vspace{-40em}

\begin{IEEEbiography}
[{\includegraphics[width=1in,height=1.25in,clip,keepaspectratio]{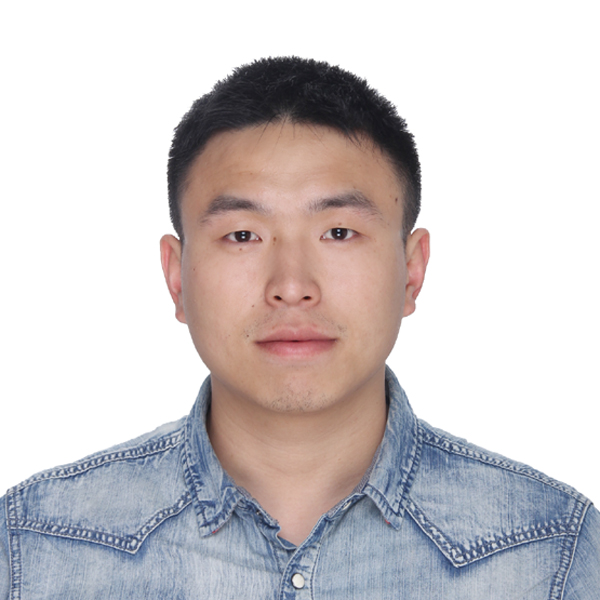}}]{Long Chen} received the Ph.D. degree in Computer Science from Zhejiang University in 2020, and the B.Eng. degree in Electrical Information Engineering from Dalian University of Technology in 2015. He is currently an assistant professor at the Department of Computer Science and Engineering, HKUST. He was a postdoctoral research scientist at Columbia University and a senior researcher at Tencent AI Lab. 
His research interests are computer vision and multimedia.
\end{IEEEbiography}

%% file: sections/V4/supplement.tex
\section*{Appendix}

\begin{table}[h!]
    \centering
    \captionsetup{labelsep=colon}
    \caption{Datasets and evaluation metrics.}
    \label{tab:datasets-evaluation-metrics}
    \resizebox{\linewidth}{!}{
    \begin{tabular}{c|c|c}
    \toprule
         \textbf{Tasks} & \textbf{\makecell[c]{Datasets \\ (Split of Base/Novel Categories)}} & \textbf{\makecell[c]{Evaluation \\ Metrics}} \\
    \midrule
    \midrule
         ZSD & \makecell[c]{Pascal VOC~\cite{pascal-voc} (16/4) \\ COCO~\cite{mscoco} (48/17, 65/15) \\ ILSVRC-2017 Detection~\cite{ilsvrc} (177/23) \\ Visual Genome~\cite{vg} (478/130)} & \makecell[c]{$\text{AP}_{50}^N$, $\text{AP}_{50}^B$, $\text{AP}_{50}$ \\ R@100} \\
    \midrule
         ZSSS & \makecell[c]{Pascal VOC~\cite{pascal-voc} (15/5) \\ Pascal Context~\cite{pascal-context} (29/4)} & \makecell[c]{$\text{mIoU}_B$, $\text{mIoU}_N$, hIoU} \\
    \midrule
         OVD & \makecell[c]{COCO~\cite{mscoco} (48/17) \\ LVIS~\cite{lvis} (866/337) \\ Objects365~\cite{objects365} \\ OpenImages~\cite{open-images-v4}} & \makecell[c]{$\text{AP}_{50}^N$, $\text{AP}_{50}^B$, $\text{AP}_{50}$, \\ $\text{AP}_r$, $\text{AP}_c$, $\text{AP}_f$, AP} \\
    \midrule
         OVSS & \makecell[c]{Pascal VOC~\cite{pascal-voc} (15/5) \\ COCO stuff~\cite{coco-stuff} (156/15) \\ ADE20K-150~\cite{ade20k} (135/15) \\ ADE20K-847~\cite{ade20k} (572/275) \\ Pascal Context-59~\cite{pascal-context} \\ Pascal Context-459~\cite{pascal-context}} & \makecell[c]{mIoU, \\ $\text{mIoU}_N$, $\text{mIoU}_B$, hIoU} \\
    \midrule
         OVIS & \makecell[c]{COCO~\cite{mscoco} (48/17) \\ ADE20k~\cite{ade20k} (135/15) \\ OpenImages~\cite{open-images-v4} (200/100)} & \makecell[c]{mask $\text{AP}_{50}$, \\ mask $\text{AP}_{50}^N$, mask $\text{AP}_{50}^B$} \\
    \midrule
         OVPS & \makecell[c]{COCO Panoptic~\cite{panoptic-segmentation} (119/14) \\ ADE20k~\cite{ade20k} \\ Cityscapes~\cite{cityscapes}} & \makecell[c]{PQ, SQ, RQ} \\
    \midrule
         OV3D & \makecell[c]{SUN RGB-D~\cite{sun-rgbd} \\ ScanNet~\cite{scannet} \\ nuScenes~\cite{nuscenes}} & \makecell[c]{$\text{AP}_{25}^N$, $\text{AP}_{25}^B$, $\text{AP}_{25}$} \\
    \midrule
         OV3SS & \makecell[c]{ScanNet~\cite{scannet} \\ nuScenes~\cite{nuscenes}} & \makecell[c]{mIoU, \\ $\text{mIoU}_N$, $\text{mIoU}_B$, hIoU} \\
    \midrule
         OV3IS & \makecell[c]{ScanNet200~\cite{scannet200}} & AP, $\text{AP}_{25}$, $\text{AP}_{50}$ \\
    \midrule
         OVVU & \makecell[c]{Youtube-VIS~\cite{ytvis} \\ BURST~\cite{burst} \\ LV-VIS~\cite{ov2seg}} & AP, $\text{AP}^B$, $\text{AP}^N$ \\   
    \bottomrule
    \end{tabular}}
\end{table}

In this supplementary material, we provide as much as detailed, comprehensive, and fair comparisons of methods for different tasks and settings. We keep track of new works at \href{https://github.com/seanzhuh/awesome-open-vocabulary-detection-and-segmentation}{awesome-ovd-ovs}. However, note that the benchmark does not differentiate the subtle nuances such as image backbone initialization weights, with or without background evaluation, and different version of validation sets, \emph{etc}. For precise details, please refer to the original paper. 

\subsection*{Evaluation Protocols, Metrics, and Datasets}
\label{subsec:evaluation-protocols-and-datasets}

ZSD and ZSS mainly use two evaluation protocols for assessment: 1) evaluating only on novel classes (\textbf{non-generalized}); 2) evaluating on both base and novel classes (\textbf{generalized}). The generalized assessment is more challenging and realistic than non-generalized evaluation. It competes novel with base classes and requires model not to overfit on base classes. OVD and OVS mainly adopt the generalized protocol for evaluation. Besides, OVD and OVS introduce the third protocol, termed cross-dataset transfer evaluation (CDTE). Namely, the model is trained on one source dataset and tested on other target datasets without adaptation. Vocabularies of source and target datasets may or may not partially overlap with each other.

The evaluation metric for object detection and instance segmentation is mainly box and mask AP at a certain IoU threshold ($\text{AP}_{50}$, $\text{AP}_{25}$) or integrated over a series of IoU threshold (0.5 to 0.95 with 0.05 as interval). The AP can be divided into $\text{AP}_B$ and $\text{AP}_N$ considering only base or novel classes. For LVIS~\cite{lvis} dataset, the rare categories are regarded as novel classes, its metric is denoted as $\text{AP}_r$, common and frequent classes are base classes. Additionally, object detection and instance segmentation use recall as a complementary metric. For semantic segmentation, the metric is mIoU only considering either base ($\text{mIoU}_{B}$) or novel ($\text{mIoU}_{N}$) classes. The hamonic mean (hIoU) between $\text{mIoU}_{B}$ and $\text{mIoU}_{N}$ is calculated as the following:

\begin{equation}
    \label{eq:hm}
    \text{hIoU}=\frac{2*\text{mIoU}_{B}*\text{mIoU}_{N}}{\text{mIoU}_{B}+\text{mIoU}_{N}}.
\end{equation}
Note that for ZSD, the $\text{AP}_{50}$ may represent the harmonic mean of $\text{AP}_{50}^{N}$ and $\text{AP}_{50}^B$ in some work, we do not differentiate them here. For panoptic segmentation, the metric is panoptic quality~\cite{panoptic-segmentation} (PQ) which can be viewed as a multiplication between segmentation quality (SQ) and recognition quality (RQ). For 3D scene and video understanding, the metrics are mainly inherited from their counterparts in image domain. For a complete dataset and metric list, \emph{c.f.} to~\cref{tab:datasets-evaluation-metrics}.

\begin{table}[h!]
    \centering
    \caption{ZSD performance on COCO~\cite{mscoco} dataset. IRv2 is InceptionResnetv2~\cite{inception}, \emph{c.f.} to~\cref{tab:zsd-pascal-voc} for other notations.}
    \resizebox{\linewidth}{!}{
    \begin{tabular}{lcc|c|c}
    \toprule
         \multirow{2}{*}{Method} & \multirow{2}{*}{\makecell[c]{Image \\ Backbone}} & \multirow{2}{*}{\makecell[c]{Semantic \\ Embeddings}} & \multirow{2}{*}{$\text{AP}_{50}^N$} & \multirow{2}{*}{$\text{AP}_{50}^B$/$\text{AP}_{50}^N$/$\text{AP}_{50}$}  \\
         & & & & \\
    \midrule
    \midrule
        \belowrulesepcolor{gray!5!}
        \rowcolor{gray!5!}\multicolumn{5}{c}{48/17 split~\cite{sb-lab}} \\
        \aboverulesepcolor{gray!5!}
    \midrule
        SAN~\cite{zsd-ijcv} & R50 & W2V & 5.1 & 13.9/2.6/4.3 \\
        SB~\cite{sb-lab} & IRv2 & - & 0.7 & - \\
        LAB~\cite{sb-lab} & IRv2 & - & 0.3 & - \\
        DSES~\cite{sb-lab} & IRv2 & - & 0.5 & - \\
        MS-Zero~\cite{mszero++} & R101 & GloVe~\cite{glove} & 12.9 & -/-/30.7 \\
        PL~\cite{polarity} & R50-FPN & W2V & 10.0 & 35.9/4.1/7.4 \\ 
        CG-ZSD~\cite{cg-zsd} & DN53~\cite{yolov3} & BERT~\cite{bert} & 7.2 & - \\ 
        BLC~\cite{blc} & RN50 & W2V & 10.6 & 42.1/4.5/8.2 \\
        ContrastZSD~\cite{contrast-zsd} & R101 & W2V & 12.5 & 45.1/6.3/11.1 \\
        SSB~\cite{ssb} & R101 & W2V & 14.8 & 48.9/10.2/16.9 \\
        DELO~\cite{delo} & DN19~\cite{yolov3} & W2VR~\cite{zsd-tcsvt} & 7.6 & -/-/13.0 \\
        RRFS~\cite{rrfs} & R101 & FT & 13.4 & 42.3/13.4/20.4 \\
    \midrule
        \belowrulesepcolor{gray!5!}
        \rowcolor{gray!5!}\multicolumn{5}{c}{65/15 split~\cite{polarity}} \\
        \aboverulesepcolor{gray!5!}
    \midrule
        PL~\cite{polarity} & R50-FPN & W2V & 12.4 & 34.1/12.4/18.2 \\
        TL~\cite{tlzsd} & R50-FPN & W2V & 14.6 & 28.8/14.1/18.9 \\
        CG-ZSD~\cite{cg-zsd} & DN53 & BERT~\cite{bert} & 10.9 & - \\
        BLC~\cite{blc} & R50 & W2V & 14.7 & 36.0/13.1/19.2 \\
        DPIF-M~\cite{dpif} & R50 & W2V & 19.8 & 29.8/19.5/23.6 \\
        ContrastZSD~\cite{contrast-zsd} & R101 & W2V & 18.6 & 40.2/16.5/23.4 \\
        SSB~\cite{ssb} & R101 & W2V & 19.6 & 40.2/19.3/26.1 \\
        SU~\cite{su} & R101 & FT & 19.0 & 36.9/19.0/25.1 \\ 
        RRFS~\cite{rrfs} & R101 & FT & 19.8 & 37.4/19.8/26.0 \\
    \bottomrule
    \end{tabular}}
    \label{tab:zsd-mscoco}
\end{table}

\begin{table}[h!]
    \centering
    \caption{ZSD performance on Pascal VOC~\cite{pascal-voc} under the non-generalized and generalized evaluation protocol. R denote ResNet~\cite{resnet}. W2V and FT is Word2Vec~\cite{word2vec} and FastText~\cite{fasttext}, respectively.}
    \resizebox{\linewidth}{!}{
    \begin{tabular}{lcc|c|c}
    \toprule
         \multirow{2}{*}{Method} & \multirow{2}{*}{\makecell[c]{Image \\ Backbone}} & \multirow{2}{*}{\makecell[c]{Semantic \\ Embeddings}} & \multirow{2}{*}{$\text{AP}_{50}^{N}$} & \multirow{2}{*}{$\text{AP}_{50}^B$/$\text{AP}_{50}^N$/$\text{AP}_{50}$} \\
         & & & & \\
    \midrule
    \midrule
        SAN~\cite{zsd-accv} & R50 & - & 59.1 & 48.0/37.0/41.8 \\
        HRE~\cite{hre} & DN19~\cite{yolov3} & aPY~\cite{apy} & 54.2 & 62.4/25.5/36.2 \\
        PL~\cite{polarity} & R50-FPN & aPY~\cite{apy} & 62.1 & - \\
        BLC~\cite{blc} & R50 & - & 55.2 & 58.2/22.9/32.9 \\
        TL~\cite{tlzsd} & R50-FPN & W2V & 66.6 & - \\
        MS-Zero~\cite{mszero++} & R101 & aPY~\cite{apy} & 62.2 & -/-/60.1 \\
        CG-ZSD~\cite{cg-zsd} & DN53~\cite{yolov3} & BERT~\cite{bert} & 54.8 & - \\
        SU~\cite{su} & R101 & FT & 64.9 & - \\
        DPIF~\cite{dpif} & R50 & aPY~\cite{apy} & - & 73.2/62.3/67.3 \\
        ContrastZSD~\cite{contrast-zsd} & R101 & aPY~\cite{apy} & 65.7 & 63.2/46.5/53.6 \\
        RRFS~\cite{rrfs} & R101 & FT & 65.5 & 47.1/49.1/48.1 \\
    \bottomrule
    \end{tabular}}
    \label{tab:zsd-pascal-voc}
\end{table}

\begin{table*}
\begin{minipage}[t!]{.5\textwidth}
\raggedleft
    \centering
    \caption{ZSD performance on ILSVRC-2017 detection~\cite{ilsvrc} and Visual Genome~\cite{vg} dataset under non-generalized evaluation protocol. R@100 is Recall@100 at IoU threshold 0.5, \emph{c.f.} to~\cref{tab:zsd-pascal-voc} and~\cref{tab:zsd-mscoco} for other notations.}
    \begin{tabular}{lcc|cc}
    \toprule
         \multirow{2}{*}{Method} & \multirow{2}{*}{\makecell[c]{Image \\ Backbone}} & \multirow{2}{*}{\makecell[c]{Semantic \\ Embeddings}} & \multirow{2}{*}{R@100} & \multirow{2}{*}{$\text{AP}_{50}^N$} \\
         & & & & \\
    \midrule
    \midrule
        \belowrulesepcolor{gray!5!}
        \rowcolor{gray!5!}\multicolumn{5}{c}{177/23 split~\cite{zsd-accv} for ILSVRC-2017 Detection} \\
        \aboverulesepcolor{gray!5!}
    \midrule
        SAN~\cite{zsd-accv} & R50 & W2V & - & 16.4 \\
        ZSDTD~\cite{zsdtd} & IRv2 & Text-Desc & - & 24.1 \\
        GTNet~\cite{gtn} & R101 & FT & - & 26.0 \\
        SU~\cite{su} & R101 & FT & - & 24.3 \\
    \midrule
        \belowrulesepcolor{gray!5!}
        \rowcolor{gray!5!}\multicolumn{5}{c}{478/130 split~\cite{zsd-accv} for Visual Genome} \\
        \aboverulesepcolor{gray!5!}
    \midrule
        SB~\cite{sb-lab} & IRv2 & - & 4.1 & - \\
        LAB~\cite{sb-lab} & IRv2 & - & 5.4 & - \\
        DESE~\cite{sb-lab} & IRv2 & - & 4.8 & - \\
        CA-ZSD~\cite{ca-zsd} & R50 & GloVe~\cite{glove} & - & - \\
        ZSDTD~\cite{zsdtd} & IRv2 & Text-Desc & 7.2 & - \\
        GTNet~\cite{gtn} & R101 & W2V & 11.3 & - \\
        S2V~\cite{s2v} & IRv2 & GloVe~\cite{glove} & 11.0 & - \\
        DPIF-M~\cite{dpif} & R50 & W2V & 18.3 & 1.8 \\
    \bottomrule
    \end{tabular}
    \label{tab:zsd-ilsvrc-vg}
\end{minipage}%
\quad
\begin{minipage}[t!]{.5\textwidth}
\raggedright
    \centering
    \caption{Zero-shot semantic segmentation performance on Pascal VOC~\cite{pascal-voc} and Pascal Context~\cite{pascal-context} datasets. ZS3Net~\cite{zs3net} randomly samples 2 to 10 novel classes with step size 2, here we only show the results of 4 novel classes. HM denote hamonic mean (hIoU), for other notations, \emph{c.f.} to~\cref{tab:zsd-pascal-voc} and~\cref{tab:zsd-mscoco}.}
    \resizebox{\linewidth}{!}{
    \begin{tabular}{lcc|c|c}
    \toprule
         \multirow{2}{*}{Method} & \multirow{2}{*}{\makecell[c]{Image \\ Backbone}} & \multirow{2}{*}{\makecell[c]{Semantic \\ Embeddings}} & Pascal VOC & Pascal Context \\
         & & & mIoU (B/N/HM) & mIoU (B/N/HM) \\
    \midrule
    \midrule
        \belowrulesepcolor{gray!5!}
        \rowcolor{gray!5!}\multicolumn{5}{c}{\makecell[c]{15/5 split~\cite{spnet} for Pascal VOC \\ 29/4 split~\cite{cagnet} for Pascal Context}} \\
        \aboverulesepcolor{gray!5!}
    \midrule    
        SPNet-C~\cite{spnet} & R101 & W2V \& FT & 78.0/15.6/26.1 & 35.1/4.0/7.2 \\
        ZS3Net~\cite{zs3net} & R101 & W2V & 77.3/17.7/28.7 & 33.0/7.7/12.5 \\
        VM~\cite{vm} & VGG16~\cite{vgg} & 
        GloVe~\cite{glove} & -/35.6/- & - \\
        CaGNet~\cite{cagnet} & R101 & W2V \& FT & 78.4/26.6/39.7 & 36.1/14.4/20.6 \\ 
        SIGN~\cite{sign} & R101 & W2V \& FT & 75.4/28.9/41.7 & 33.7/14.9/20.7 \\
    \midrule
        \belowrulesepcolor{gray!5!}
        \rowcolor{gray!5!}\multicolumn{5}{c}{Novel - 4~\cite{zs3net}} \\
        \aboverulesepcolor{gray!5!}
    \midrule
        SPNet~\cite{spnet} & R101 & W2V \& FT & 67.3/21.8/32.9 & 36.3/18.1/24.2 \\
        ZS3Net~\cite{zs3net} & R101 & W2V & 66.4/23.2/34.4 & 37.2/24.9/29.8 \\
        CSRL~\cite{csrl} & R101 & - & 69.8/31.7/43.6 & 39.8/23.9/29.9 \\
        JoEm~\cite{joem} & R101 & W2V & 67.0/33.4/44.6 & 36.9/30.7/33.5 \\
        PMOSR~\cite{pmosr} & R101 & W2V & 75.0/44.1/55.5 & 41.1/43.1/42.1 \\
    \bottomrule
    \end{tabular}}
    \label{tab:zsss-pascal-voc-context}
\end{minipage}
\end{table*}

\begin{table*}
\centering
    \caption{Open-vocabulary 3D detection performance on SUN RGB-D~\cite{sun-rgbd}, ScanNet~\cite{scannet}, and nuScenes~\cite{nuscenes} datasets under the generalized evaluation and CDTE protocol.}
    \begin{tabular}{lccc|ccc|ccc|c}
    \toprule
         \multirow{2}{*}{Method} & \multirow{2}{*}{Detector} & \multirow{2}{*}{\makecell[c]{3D \\ Annotations}} & \multirow{2}{*}{\makecell[c]{2D \\ Detector}} & \multicolumn{3}{c}{SUN RGB-D} \vline & \multicolumn{3}{c}{ScanNet} \vline & nuScenes \\
         & & & & $\text{AP}_{25}^N$ & $\text{AP}_{25}^B$ & $\text{AP}_{25}$ & $\text{AP}_{25}^N$ & $\text{AP}_{25}^N$ & $\text{AP}_{25}$ & $\text{AP}_{25}$ \\
    \midrule
    \midrule
        OV-3DET~\cite{ov-3det} & 3DETR~\cite{3detr} & \XSolidBrush & Detic~\cite{detic} & - & - & 20.5 & - & - & 18.0 & - \\
        FM-OV3D~\cite{fm-ov3d} & 3DETR~\cite{3detr} & \Checkmark & Grounded-SAM~\cite{grounded-sam} & - & - & 21.5 & - & - & 21.5 & - \\
        OpenSight~\cite{opensight} & VoxelNet~\cite{voxelnet} & \XSolidBrush & Grounding DINO~\cite{grounding-dino} & - & - & - & - & - & - & 23.5 \\
        CoDA~\cite{coda} & 3DETR~\cite{3detr} & \Checkmark & \XSolidBrush & 6.7 & 38.7 & 13.7 & 6.5 & 21.6 & 9.0 & - \\
        L3Det~\cite{object2scene} & L3Det~\cite{object2scene} & \Checkmark & \XSolidBrush & 24.6 & - & - & 25.4 & - & - & - \\
    \bottomrule
    \end{tabular}
    \label{tab:ov3d-det}
\end{table*}

\begin{table*}[t!]
\centering
    \caption{Open-vocabulary 3D instance segmentation performance on ScanNet200~\cite{scannet200} dataset.}
    \begin{tabular}{lccc|ccc}
    \toprule
         \multirow{2}{*}{Method} & \multirow{2}{*}{Segmentor} & \multirow{2}{*}{\makecell[c]{3D \\ Annotations}} & \multirow{2}{*}{\makecell[c]{2D \\ Segmentor}} & \multicolumn{3}{c}{ScanNet200} \\
         & & & & AP & $\text{AP}_{25}$ & $\text{AP}_{50}$ \\
    \midrule
    \midrule
        OpenMask3D~\cite{openmask3d} & Mask3D~\cite{mask3d} & \Checkmark & SAM~\cite{sam} & 12.8 & 19.0 & 16.8 \\
        MaskClustering~\cite{mask-clustering} & - & \XSolidBrush & CropFormer~\cite{cropformer} & 12.0 & 30.1 & 23.3 \\
        Open3DIS~\cite{open3dis} & Mask3D~\cite{mask3d} & \Checkmark & Grounded-SAM~\cite{grounded-sam} & 23.7 & 32.8 & 29.4 \\
    \bottomrule
    \end{tabular}
    \label{tab:ov3d-ins-seg}
\end{table*}

\begin{table*}[t!]
\centering
    \caption{Open-vocabulary video instance segmentation performance on validation set of Youtube-VIS19~\cite{ytvis} (YTVIS-19), Youtube-VIS21~\cite{ytvis} (YTVIS-21), BURST~\cite{burst}, and LV-VIS~\cite{ov2seg}.}
    \resizebox{\textwidth}{!}{
    \begin{tabular}{lcccc|ccc|ccc|ccc|ccc}
    \toprule
         \multirow{2}{*}{Method} & \multirow{2}{*}{Segmentor} & \multirow{2}{*}{Tracker} & \multirow{2}{*}{\makecell[c]{Training \\ Source}} & \multirow{2}{*}{Prompts} & \multicolumn{3}{c}{YTVIS-19} \vline & \multicolumn{3}{c}{YTVIS-21} \vline & \multicolumn{3}{c}{BURST} \vline & \multicolumn{3}{c}{LV-VIS} \\
         & & & & & AP & $\text{AP}^B$ & $\text{AP}^N$ & AP & $\text{AP}^B$ & $\text{AP}^N$ & AP & $\text{AP}^B$ & $\text{AP}^N$ & AP & $\text{AP}^B$ & $\text{AP}^N$ \\
    \midrule
    \midrule
        OV2Seg~\cite{ov2seg} & - & SORT~\cite{sort} & LVIS~\cite{lvis} & L (cat) & 37.6 & 41.1 & 21.3 & 33.9 & 36.7 & 18.2 & 4.9 & 5.3 & 3.0 & 21.1 & 27.5 & 16.3 \\
        OpenVIS~\cite{openvis} & M2F~\cite{mask2former} & \XSolidBrush & YTVIS~\cite{ytvis} & T (cat) & - & - & - & - & - & - & 3.5 & 5.8 & 3.0 & - & - & - \\
        BriVIS~\cite{brivis} & M2F~\cite{mask2former} & \XSolidBrush & LV-VIS & T (cat) & 45.3 & - & - & 39.5 & - & - & 7.4 & 9.5 & 6.9 & 27.68 & - & - \\
    \bottomrule
    \end{tabular}}
    \label{tab:ov3d-vid-ins-seg}
\end{table*}

\begin{table*}[t]
    \centering
    \caption{OVD performance on COCO~\cite{mscoco} under generalized evaluation protocol. ``T'' and ``L'' denote template and learnable prompts. ``cat'' and ``desc'' denote that the prompts are filled with class names or class descriptions (definitions, synonyms, \emph{etc}). ``Ensemble'' represents that whether detector prediction is ensembled with CLIP prediction~\cite{f-vlm,vild} or not. COCO Cap is COCO Captions dataset~\cite{coco-captions}, Visual Genome~\cite{vg} is denoted as VG, Conceptual Captions~\cite{cc} is denoted as CC3M.}
    \resizebox{\linewidth}{!}{
    \begin{tabular}{lcccccc|ccc}
    \toprule
         \multirow{2}{*}{Method} & \multirow{2}{*}{\makecell[c]{Image \\ Backbone}} & \multirow{2}{*}{Detector} & \multirow{2}{*}{\makecell[c]{Image-Text \\ Pairs}} & \multirow{2}{*}{\makecell[c]{Text \\ Encoder}} & \multirow{2}{*}{Prompts} & \multirow{2}{*}{Ensemble} & \multirow{2}{*}{$\text{AP}_{50}^{N}$} & \multirow{2}{*}{$\text{AP}_{50}^{B}$} & \multirow{2}{*}{$\text{AP}_{50}$} \\
         & & & & & & & & & \\
    \midrule
    \midrule
        \belowrulesepcolor{gray!5!}
        \rowcolor{gray!5!}\multicolumn{10}{c}{\textbf{Region-Aware Training}} \\
        \aboverulesepcolor{gray!5!}
    \midrule
        OVR-CNN~\cite{ovr-cnn} & R50 & FRCNN & COCO Cap & BERT & \XSolidBrush & \XSolidBrush & 22.8 & 46.0 & 39.9 \\ 
        LocOv~\cite{locov} & R50 & FRCNN & COCO Cap & BERT & \XSolidBrush & \XSolidBrush & 28.6 & 51.3 & 45.7 \\
        MMC-Det~\cite{mmc-det} & R50 & FRCNN & COCO Cap & BERT & \XSolidBrush & \XSolidBrush & 33.5 & - & 47.5 \\
        WSOVOD~\cite{wsovod} & DRN~\cite{drn} & FRCNN & \XSolidBrush & CLIP & T (cat) & \XSolidBrush & 35.0 & 27.9 & 29.8 \\
        RO-ViT~\cite{ro-vit} & ViT-B/16 & MRCNN & ALIGN~\cite{align} & CLIP & T (cat) & \Checkmark & 30.2 & - & 41.5 \\ 
        CFM-ViT~\cite{cfm-vit} & ViT-B/16 & MRCNN & ALIGN~\cite{align} & CLIP & T (cat) & \Checkmark & 30.8 & - & 42.4 \\ 
        DITO~\cite{dito} & ViT-B/16 & FRCNN & ALIGN~\cite{align} & CLIP & T (cat) & \Checkmark & 38.6 & - & 48.5 \\
        VLDet~\cite{vldet} & R50 & FRCNN & COCO Cap & CLIP & T (cat) & \XSolidBrush & 32.0 & 50.6 & 45.8 \\
        GOAT~\cite{goat} & R50 & FRCNN & COCO Cap & CLIP & T (cat) & \XSolidBrush & 31.7 & 51.3 & 45.7 \\
        OV-DETR~\cite{ov-detr} & R50 & Def-DETR~\cite{deformable-detr} & \XSolidBrush & CLIP & T (cat) & \XSolidBrush & 29.4 & 61.0 & 52.7 \\
        Prompt-OVD~\cite{prompt-ovd} & ViT-B/16 & Def-DETR~\cite{deformable-detr} & \XSolidBrush & CLIP & T (cat) & \Checkmark & 30.6 & 63.5 & 54.9 \\ 
        CORA~\cite{cora} & R50 & SAM-DETR~\cite{sam-detr} & \XSolidBrush & CLIP & T (cat) & \XSolidBrush & 35.1 & 35.5 & 35.4 \\
        EdaDet~\cite{edadet} & R50 & Def-DETR~\cite{deformable-detr} & CLIP~\cite{clip} & CLIP & T (cat) & \Checkmark & 35.1 & 35.5 & 35.4 \\
        SGDN~\cite{sgdn} & R50 & Def-DETR~\cite{deformable-detr} & VG, Flickr30K~\cite{flickr30k} & RoBERTa~\cite{roberta} & \XSolidBrush & \XSolidBrush & 37.5 & 61.0 & 54.9 \\
    \midrule
        \belowrulesepcolor{gray!5!}
        \rowcolor{gray!5!}\multicolumn{10}{c}{\textbf{Pseudo-Labeling}} \\
        \aboverulesepcolor{gray!5!}
    \midrule
        RegionCLIP~\cite{regionclip} & R50 & FRCNN & CC3M & CLIP & T (cat) & \XSolidBrush & 31.4 & 57.1 & 50.4 \\
        CondHead~\cite{condhead} & R50 & RegionCLIP~\cite{regionclip} & \XSolidBrush & CLIP & T (cat) & \XSolidBrush & 33.7 & 58.0 & 51.7 \\
        VL-PLM~\cite{vl-plm} & R50 & FRCNN & \XSolidBrush & CLIP & T (cat) & \XSolidBrush & 34.4 & 60.2 & 53.5 \\
        PromptDet~\cite{prompt-det} & R50 & MRCNN & LAION~\cite{laion-400m} & CLIP & L (cat+desc) & \XSolidBrush & 26.6 & - & 50.6 \\ 
        SAS-Det~\cite{sas-det} & R50 & FRCNN & \XSolidBrush & CLIP & T (cat) & \Checkmark & 37.4 & 58.5 & 53.0 \\
    \noalign{\smallskip}
    \hline
    \noalign{\smallskip}
        \multirow{2}{*}{PB-OVD~\cite{pb-ovd}} & \multirow{2}{*}{R50} & \multirow{2}{*}{MRCNN} & \multirow{2}{*}{\makecell[c]{COCO Cap, VG, \\ SBU~\cite{sbu}}} & \multirow{2}{*}{CLIP} & \multirow{2}{*}{T (cat)} & \multirow{2}{*}{\XSolidBrush} & \multirow{2}{*}{30.8} & \multirow{2}{*}{46.1} & \multirow{2}{*}{42.1} \\
        & & & & & & & & & \\
    \noalign{\smallskip}
    \hline
    \noalign{\smallskip}
        CLIM~\cite{clim} & R50 & Detic~\cite{detic} & COCO Cap & CLIP & T (cat) & \XSolidBrush & 35.4 & - & - \\
        VTP-OVD~\cite{vtp-ovd} & R50 & MRCNN & \XSolidBrush & CLIP & T (cat) & \XSolidBrush & 31.5 & 51.9 & 46.6 \\
        ProxyDet~\cite{proxy-det} & R50 & FRCNN & COCO Cap & CLIP & T (cat) & \Checkmark & 30.4 & 52.6 & 46.8 \\
        CoDet~\cite{codet} & R50 & FRCNN & COCO Cap & CLIP & T (cat) & \Checkmark & 30.6 & 52.3 & 46.6 \\
        Detic~\cite{detic} & R50 & FRCNN & COCO Cap & CLIP & T (cat) & \XSolidBrush & 27.8 & 47.1 & 45.0 \\
    \midrule
        \belowrulesepcolor{gray!5!}
        \rowcolor{gray!5!}\multicolumn{10}{c}{\textbf{Knowledge Distillation}} \\
        \aboverulesepcolor{gray!5!}
    \midrule
        ViLD~\cite{vild} & R50 & MRCNN & \XSolidBrush & CLIP & T (cat) & \Checkmark & 27.6 & 59.5 & 51.3 \\
        ZSD-YOLO~\cite{zsd-yolo} & CSP~\cite{cspnet} & YOLOv5x~\cite{yolov5} & \XSolidBrush & CLIP & T (cat+desc) & \XSolidBrush & 13.6 & 31.7 & 19.0 \\
        LP-OVOD~\cite{lp-ovod} & R50 & FRCNN & \XSolidBrush & CLIP & T (cat) & \Checkmark & 40.5 & 60.5 & 55.2 \\
        EZSD~\cite{ezsd} & R50 & MRCNN & \XSolidBrush & CLIP & T (cat) & \XSolidBrush & 31.6 & 59.9 & 52.1 \\
        SIC-CADS~\cite{sic-cads} & R50 & BARON~\cite{baron} & \XSolidBrush & CLIP & T (cat) & \Checkmark & 36.9 & 56.1 & 51.1 \\
        BARON~\cite{baron} & R50 & FRCNN & COCO Cap & CLIP & T (cat) & \XSolidBrush & 42.7 & 54.9 & 51.7 \\
        OADP~\cite{oadp} & R50 & FRCNN & \XSolidBrush & CLIP & T (cat) & \Checkmark & 35.6 & 55.8 & 50.5 \\
        RKDWTF~\cite{rkdwtf} & R50 & FRCNN & COCO Cap & CLIP & T (cat) & \XSolidBrush & 36.6 & 54.0 & 49.4 \\
        DK-DETR~\cite{dk-detr} & R50 & Def-DETR~\cite{deformable-detr} & \XSolidBrush & CLIP & T (cat) & \XSolidBrush & 32.3 & 61.1 & - \\
        HierKD~\cite{hierkd} & R50 & ATSS~\cite{atss} & CC3M & CLIP & T (cat/desc) & \XSolidBrush & 20.3 & 51.3 & 43.2 \\
        CLIPSelf~\cite{clipself} & ViT-B/16 & F-VLM~\cite{f-vlm} & \XSolidBrush & CLIP & T (cat) & \Checkmark & 37.6 & - & - \\
    \midrule
        \belowrulesepcolor{gray!5!}
        \rowcolor{gray!5!}\multicolumn{10}{c}{\textbf{Transfer Learning}} \\
        \aboverulesepcolor{gray!5!}
    \midrule
        F-VLM~\cite{f-vlm} & R50 & MRCNN & \XSolidBrush & CLIP & T (cat) & \Checkmark & 28.0 & - & 39.6 \\
        DRR~\cite{drr} & R50 & FRCNN & CC3M & CLIP & T (cat) & \XSolidBrush & 35.8 & 54.6 & 49.6 \\
    \bottomrule
    \end{tabular}}
    \label{tab:ovd-mscoco}
\end{table*}

\begin{table*}[t!]
    \centering
    \caption{OVD performance on LVIS~\cite{lvis} dataset under generalized evaluation protocol. Base classes are common and frequent classes in LVIS, rare classes in LVIS are novel class. \textcolor{gray}{Gray} numbers denote mask AP. Conceptual 12M dataset~\cite{cc12m} is denoted as CC12M. The subset of ImageNet-21K~\cite{ilsvrc} (IN21K) that overlaps with LVIS vocabulary is IN-L~\cite{detic} (\emph{c.f.} to~\cref{tab:ovd-mscoco} for other abbreviatios).}
    \resizebox{\linewidth}{!}{
    \begin{tabular}{lcccccc|cccc}
    \toprule
         \multirow{2}{*}{Method} & \multirow{2}{*}{\makecell[c]{Image \\ Backbone}} & \multirow{2}{*}{Detector} & \multirow{2}{*}{\makecell[c]{Image-Text \\ Pairs}} & \multirow{2}{*}{\makecell[c]{Text \\ Encoder}} & \multirow{2}{*}{Prompts} & \multirow{2}{*}{Ensemble} & \multirow{2}{*}{$\text{AP}_r$} & \multirow{2}{*}{$\text{AP}_c$} & \multirow{2}{*}{$\text{AP}_f$} & \multirow{2}{*}{$\text{AP}$} \\
         & & & & & & & & \\
    \midrule
    \midrule
        \belowrulesepcolor{gray!5!}
        \rowcolor{gray!5!}\multicolumn{11}{c}{\textbf{Region-Aware Training}} \\
        \aboverulesepcolor{gray!5!}
    \midrule
        MMC-Det~\cite{mmc-det} & R50 & CN2~\cite{centernetv2} & CC3M & CLIP & T (cat) & \XSolidBrush & \textcolor{gray}{21.1} & \textcolor{gray}{30.9} & \textcolor{gray}{35.5} & \textcolor{gray}{31.0} \\
        RO-ViT~\cite{ro-vit} & ViT-B/16 & MRCNN & ALIGN~\cite{align} & CLIP & T (cat) & \Checkmark & \textcolor{gray}{28.0} & - & - & \textcolor{gray}{30.2} \\
        CFM-ViT~\cite{cfm-vit} & ViT-B/16 & MRCNN & ALIGN~\cite{align} & CLIP & T (cat) & \Checkmark & $\text{29.6}_{\text{\textcolor{gray}{28.8}}}$ & - & - & $\text{33.8}_{\text{\textcolor{gray}{32.0}}}$ \\
        DITO~\cite{dito} & ViT-B/16 & FRCNN & ALIGN~\cite{align} & CLIP & T (cat) & \Checkmark & $\text{34.9}_{\text{\textcolor{gray}{32.5}}}$ & - & - & $\text{36.9}_{\text{\textcolor{gray}{34.0}}}$ \\ 
        VLDet~\cite{vldet} & R50 & CN2~\cite{centernetv2} & CC3M & CLIP & T (cat) & \XSolidBrush & \textcolor{gray}{21.7} & \textcolor{gray}{29.8} & \textcolor{gray}{34.3} & \textcolor{gray}{30.1} \\
        GOAT~\cite{goat} & R50 & CN2~\cite{centernetv2} & CC3M & CLIP & T (cat) & \XSolidBrush & \textcolor{gray}{23.3} & \textcolor{gray}{29.7} & \textcolor{gray}{34.3} & \textcolor{gray}{30.4} \\ 
        OV-DETR~\cite{ov-detr} & R50 & Def-DETR~\cite{deformable-detr} & \XSolidBrush & CLIP & T (cat) & \XSolidBrush & $\text{\textcolor{gray}{17.4}}$ & \textcolor{gray}{25.0} & \textcolor{gray}{32.5} & $\text{\textcolor{gray}{26.6}}$ \\
        Prompt-OVD~\cite{prompt-ovd} & ViT-B/16 & Def-DETR~\cite{deformable-detr} & \XSolidBrush & CLIP & T (cat) & \XSolidBrush & $\text{29.4}_{\text{\textcolor{gray}{23.1}}}$ & - & - & $\text{33.0}_{\text{\textcolor{gray}{24.2}}}$ \\ 
        CORA~\cite{cora} & R50x4 & CN2~\cite{centernetv2} & \XSolidBrush & CLIP & T (cat) & \XSolidBrush & 28.1 & - & - & - \\
        EdaDet~\cite{edadet} & R50 & Def-DETR~\cite{deformable-detr} & \XSolidBrush & CLIP & T (cat) & \Checkmark & \textcolor{gray}{23.7} & \textcolor{gray}{27.5} & \textcolor{gray}{29.1} & \textcolor{gray}{27.5} \\
        \multirow{2}{*}{SGDN~\cite{sgdn}} & \multirow{2}{*}{R50} & \multirow{2}{*}{Def-DETR~\cite{deformable-detr}} & \multirow{2}{*}{\makecell[c]{VG, \\ Flickr30K~\cite{flickr30k}}} & \multirow{2}{*}{RoBERTa~\cite{roberta}} & \multirow{2}{*}{\XSolidBrush} & \multirow{2}{*}{\XSolidBrush} & \multirow{2}{*}{\textcolor{gray}{23.6}} & \multirow{2}{*}{\textcolor{gray}{29.0}} & \multirow{2}{*}{\textcolor{gray}{34.3}} & \multirow{2}{*}{\textcolor{gray}{31.1}} \\
        & & & & & & & & & & \\
    \midrule
        \belowrulesepcolor{gray!5!}
        \rowcolor{gray!5!}\multicolumn{11}{c}{\textbf{Pseudo-Labeling}} \\
        \aboverulesepcolor{gray!5!}
    \midrule
        RegionCLIP~\cite{regionclip} & R50 & MRCNN & CC3M & CLIP & T (cat) & \XSolidBrush & $\text{17.1}_{\text{\textcolor{gray}{17.4}}}$ & $\text{27.4}_{\text{\textcolor{gray}{26.0}}}$ & $\text{34.0}_{\text{\textcolor{gray}{31.6}}}$ & $\text{28.2}_{\text{\textcolor{gray}{26.7}}}$ \\
        CondHead~\cite{condhead} & R50 & RegionCLIP~\cite{regionclip} & \XSolidBrush & CLIP & T (cat) & \XSolidBrush & $\text{19.9}_{\text{\textcolor{gray}{20.0}}}$ & $\text{28.6}_{\text{\textcolor{gray}{27.3}}}$ & $\text{35.2}_{\text{\textcolor{gray}{32.2}}}$ & $\text{29.7}_{\text{\textcolor{gray}{27.9}}}$ \\
        PromptDet~\cite{prompt-det} & R50 & MRCNN & LAION~\cite{laion-400m} & CLIP & L (cat+desc) & \XSolidBrush & \textcolor{gray}{21.4} & \textcolor{gray}{23.3} & \textcolor{gray}{29.3} & \textcolor{gray}{25.3} \\ 
        SAS-Det~\cite{sas-det} & R50 & FRCNN & \XSolidBrush & CLIP & T (cat) & \Checkmark & 20.9 & 26.1 & 31.6 & 27.4 \\
        CLIM~\cite{clim} & R50 & VLDet~\cite{detic} & CC3M & CLIP & T (cat) & \XSolidBrush & \textcolor{gray}{22.2} & - & - & - \\
        ProxyDet~\cite{proxy-det} & R50 & CN2~\cite{centernetv2} & IN-L & CLIP & T (cat) & \Checkmark & \textcolor{gray}{26.2} & - & - & \textcolor{gray}{32.5} \\
        CoDet~\cite{codet} & R50 & CN2~\cite{centernetv2} & CC3M & CLIP & T (cat) & \XSolidBrush & \textcolor{gray}{23.4} & \textcolor{gray}{30.0} & \textcolor{gray}{34.6} & \textcolor{gray}{30.7} \\ 
        Detic~\cite{detic} & R50 & CN2~\cite{centernetv2} & IN-L & CLIP & T (cat) & \XSolidBrush & \textcolor{gray}{24.6} & - & - & \textcolor{gray}{32.4} \\ 
        MMC~\cite{mmc} & R50 & CN2~\cite{centernetv2} & IN-L & CLIP & GPT-3~\cite{gpt-3} & \XSolidBrush & \textcolor{gray}{27.3} & - & - & \textcolor{gray}{33.1} \\ 
        3Ways~\cite{3ways} & NF-F0~\cite{nfnet} & FCOS~\cite{fcos} & CC12M & CLIP & T (cat) & \XSolidBrush & 25.6 & 34.2 & 41.8 & 35.7 \\
        PLAC~\cite{plac} & Swin-B & Def-DETR~\cite{deformable-detr} & CC3M & CLIP & T (cat) & \XSolidBrush & 27.0 & 40.0 & 44.5 & 39.5 \\
    \midrule
        \belowrulesepcolor{gray!5!}
        \rowcolor{gray!5!}\multicolumn{11}{c}{\textbf{Knowledge Distillation}} \\
        \aboverulesepcolor{gray!5!}
    \midrule
        ViLD-ens~\cite{vild} & R50 & MRCNN & \XSolidBrush & CLIP & T (cat) & \Checkmark & $\text{16.7}_{\text{\textcolor{gray}{16.6}}}$ & $\text{26.5}_{\text{\textcolor{gray}{24.6}}}$ & $\text{34.2}_{\text{\textcolor{gray}{30.3}}}$ & $\text{27.8}_{\text{\textcolor{gray}{25.5}}}$ \\
        LP-OVOD~\cite{lp-ovod} & R50 & MRCNN & \XSolidBrush & CLIP & T (cat) & \Checkmark & \textcolor{gray}{19.3} & \textcolor{gray}{26.1} & \textcolor{gray}{29.4} & \textcolor{gray}{26.2} \\
        EZSD~\cite{ezsd} & R50 & MRCNN & \XSolidBrush & CLIP & T (cat) & \XSolidBrush & 15.8 & 25.6 & 31.7 & 26.3 \\
        SIC-CADS~\cite{sic-cads} & R50 & Detic~\cite{detic} & IN21K & CLIP & T (cat) & \Checkmark & \textcolor{gray}{26.5} & \textcolor{gray}{33.0} & \textcolor{gray}{35.6} & \textcolor{gray}{32.9} \\ 
        BARON~\cite{baron} & R50 & FRCNN & \XSolidBrush & CLIP & L (cat) & \XSolidBrush & $\text{23.2}_{\text{\textcolor{gray}{22.6}}}$ & $\text{29.3}_{\text{\textcolor{gray}{27.6}}}$ & $\text{32.5}_{\text{\textcolor{gray}{29.8}}}$ & 
        $\text{29.5}_{\text{\textcolor{gray}{27.6}}}$ \\
        OADP~\cite{oadp} & R50 & FRCNN & \XSolidBrush & CLIP & T (cat) & \Checkmark & $\text{21.9}_{\text{\textcolor{gray}{21.7}}}$ & $\text{28.4}_{\text{\textcolor{gray}{26.3}}}$ & $\text{32.0}_{\text{\textcolor{gray}{29.0}}}$ & $\text{28.7}_{\text{\textcolor{gray}{26.6}}}$ \\
        GridCLIP~\cite{grid-clip} & R50 & FCOS~\cite{fcos} & \XSolidBrush & CLIP & T (cat) & \XSolidBrush & 15.0 & 22.7 & 32.5 & 25.2 \\
        RKDWTF~\cite{rkdwtf} & R50 & CN2~\cite{centernetv2} & IN21K & CLIP & T (cat) & \XSolidBrush & \textcolor{gray}{25.2} & \textcolor{gray}{33.4} & \textcolor{gray}{35.8} & \textcolor{gray}{32.9} \\
        DK-DETR~\cite{dk-detr} & R50 & Def-DETR~\cite{deformable-detr} & \XSolidBrush & CLIP & T (cat) & \XSolidBrush & $\text{22.2}_{\text{\textcolor{gray}{20.5}}}$ & $\text{32.0}_{\text{\textcolor{gray}{28.9}}}$ & $\text{40.2}_{\text{\textcolor{gray}{35.4}}}$ & 
        $\text{33.5}_{\text{\textcolor{gray}{30.0}}}$ \\ 
        DetPro~\cite{detpro} & R50 & MRCNN & \XSolidBrush & CLIP & L (cat) & \Checkmark & $\text{20.8}_{\text{\textcolor{gray}{19.8}}}$ & $\text{27.8}_{\text{\textcolor{gray}{25.6}}}$ & $\text{32.4}_{\text{\textcolor{gray}{28.9}}}$ & $\text{28.4}_{\text{\textcolor{gray}{25.9}}}$ \\
        CLIPSelf~\cite{clipseg} & ViT-B/16 & F-VLM~\cite{f-vlm} & \XSolidBrush & CLIP & T (cat) & \Checkmark & \textcolor{gray}{25.3} & - & - & - \\
    \midrule
        \belowrulesepcolor{gray!5!}
        \rowcolor{gray!5!}\multicolumn{11}{c}{\textbf{Transfer Learning}} \\
        \aboverulesepcolor{gray!5!}
    \midrule
        OWL-ViT~\cite{owl-vit} & ViT-H/14 & DETR & LiT~\cite{lit} & CLIP & T (cat) & \XSolidBrush & 23.3 & - & - & 35.3 \\
        F-VLM~\cite{f-vlm} & R50 & MRCNN & \XSolidBrush & CLIP & T (cat) & \Checkmark & \textcolor{gray}{18.6} & - & - & \textcolor{gray}{24.2} \\
    \bottomrule
    \end{tabular}}
    \label{tab:ovd-lvis}
\end{table*}

\begin{table*}
\centering
    \caption{OVD performance under the CDTE protocol on Pascal VOC~\cite{pascal-voc} (VOC), Obejcts365~\cite{objects365} (O365), COCO~\cite{mscoco}, OpenImages~\cite{open-images-v4} (OI), and LVIS~\cite{lvis} validation sets. Cap4M image-text pairs are crawled in~\cite{glip}. GoldG denote the merged grounding datasets in~\cite{mdetr,glip} (\emph{c.f.} to~\cref{tab:ovd-lvis} for other notations). Note that some methods evaluate on different versions of O365 and OI validation datasets, we do not differentiate them here. \textcolor{gray}{Gray} Numbers denote the performance of LVIS \emph{minival} set~\cite{mdetr}. All metrics are box AP.}
    \resizebox{\linewidth}{!}{
    \begin{tabular}{lccc|c|cc|cc|c|c}
    \toprule
         \multirow{2}{*}{Method} & \multirow{2}{*}{\makecell[c]{Image \\ Backbone}} & \multirow{2}{*}{Detector} & \multirow{2}{*}{\makecell[c]{Training Source}} & VOC & \multicolumn{2}{c}{COCO} \vline & \multicolumn{2}{c}{O365} \vline & OI & LVIS \\
         & & & & $\text{AP}_{50}$ & AP & $\text{AP}_{50}$ & AP & $\text{AP}_{50}$ & $\text{AP}_{50}$ & $\text{AP}_{r}$/AP \\
    \midrule
    \midrule
        \belowrulesepcolor{gray!5!}
        \rowcolor{gray!5!}\multicolumn{11}{c}{\textbf{Region-Aware Training}} \\
        \aboverulesepcolor{gray!5!}
    \midrule
        MMC-Det~\cite{mmc-det} & R50 & CN2~\cite{centernetv2} & LVIS, CC3M & - & - & 56.4 & - & 21.4 & 38.6 & - \\ 
    \noalign{\smallskip}
    \hline
    \noalign{\smallskip}
        \multirow{2}{*}{DetCLIP~\cite{detclip}} & \multirow{2}{*}{Swin-T} & \multirow{2}{*}{ATSS~\cite{atss}} & \multirow{2}{*}{\makecell[c]{O365, GoldG, \\ YFCC1M~\cite{yfcc100m}}} & \multirow{2}{*}{-} & \multirow{2}{*}{-} & \multirow{2}{*}{-} & \multirow{2}{*}{-} & \multirow{2}{*}{-} & \multirow{2}{*}{-} & \multirow{2}{*}{${25.0}_{\textcolor{gray}{33.2}}$/${28.4}_{\textcolor{gray}{35.9}}$} \\
        & & & & & & & & & & \\
    \noalign{\smallskip}
    \hline
    \noalign{\smallskip}
        \multirow{2}{*}{DetCLIPv2~\cite{detclipv2}} & \multirow{2}{*}{Swin-T} & \multirow{2}{*}{ATSS~\cite{atss}} & \multirow{2}{*}{\makecell[c]{O365, GoldG, \\ CC3M, CC12M}} & \multirow{2}{*}{-} & \multirow{2}{*}{-} & \multirow{2}{*}{-} & \multirow{2}{*}{-} & \multirow{2}{*}{-} & \multirow{2}{*}{-} & \multirow{2}{*}{\textcolor{gray}{36.0}/\textcolor{gray}{40.4}} \\
        & & & & & & & & & & \\
    \noalign{\smallskip}
    \hline
    \noalign{\smallskip}
        RO-ViT~\cite{ro-vit} & ViT-B/16 & MRCNN & LVIS, ALIGN~\cite{align} & - & - & - & 17.1 & 26.9 & - & - \\
        CFM-ViT~\cite{cfm-vit} & ViT-B/16 & MRCNN & LVIS, ALIGN~\cite{align} & - & - & - & 15.9 & 24.6 & - & - \\
        DITO~\cite{dito} & ViT-L/16 & FRCNN & LVIS, ALIGN~\cite{align} & - & - & - & 19.8 & 30.4 & - & - \\
        OV-DETR~\cite{ov-detr} & R50 & Def-DETR~\cite{deformable-detr} & LVIS & 76.1 & 38.1 & 58.4 & - & - & - & - \\
        EdaDet~\cite{edadet} & R50 & Def-DETR~\cite{deformable-detr} & LVIS & - & - & - & 13.6 & 19.8 & - & - \\
        MDETR~\cite{mdetr} & R101 & DETR & GoldG+~\cite{mdetr} & - & - & - & - & - & - & ${7.4}_{\textcolor{gray}{20.9}}$/${22.5}_{\textcolor{gray}{24.2}}$ \\
        MQ-Det~\cite{mq-det} & Swin-T & GLIP~\cite{glip} & O365 & - & - & - & - & - & - & ${15.4}_{\textcolor{gray}{21.0}}$/${22.6}_{\textcolor{gray}{30.4}}$ \\
        YOLO-World~\cite{yolo-world} & - & YOLOv8-L & O365, GoldG & - & - & - & - & - & - & \textcolor{gray}{27.1}/\textcolor{gray}{35.0} \\
        SGDN~\cite{sgdn} & R50 & Def-DETR~\cite{deformable-detr} & LVIS, Flickr30K, VG & - & 40.5 & - & - & - & - & - \\
    \midrule
        \belowrulesepcolor{gray!5!}
        \rowcolor{gray!5!}\multicolumn{11}{c}{\textbf{Pseudo-Labeling}} \\
        \aboverulesepcolor{gray!5!}
    \midrule
        GLIP~\cite{glip} & Swin-T & DyHead~\cite{dyhead} & O365, GoldG, Cap4M & - & 46.3 & - & - & - & - & ${10.1}_{\textcolor{gray}{20.8}}$/$17.2_{\textcolor{gray}{26.0}}$ \\
        GLIPv2~\cite{glipv2} & Swin-T & DyHead~\cite{dyhead} & O365, GoldG, Cap4M & - & - & - & - & - & - & -/\textcolor{gray}{29.0} \\
        Grounding DINO~\cite{grounding-dino} & Swin-T & DINO~\cite{dino-detr} & O365, GoldG, Cap4M & - & 48.4 & - & - & - & - & \textcolor{gray}{18.1}/\textcolor{gray}{27.4} \\
    \noalign{\smallskip}
    \hline
    \noalign{\smallskip}
        \multirow{2}{*}{PB-OVD~\cite{pb-ovd}} & \multirow{2}{*}{R50} & \multirow{2}{*}{MRCNN} & COCO, COCO Cap & \multirow{2}{*}{59.2} & \multirow{2}{*}{-} & \multirow{2}{*}{-} & \multirow{2}{*}{6.9} & \multirow{2}{*}{-} & \multirow{2}{*}{-} & \multirow{2}{*}{-} \\
        & & & VG, SBU~\cite{sbu} & & & & & & & \\
    \noalign{\smallskip}
    \hline
    \noalign{\smallskip}
        VTP-OVD~\cite{vtp-ovd} & R50 & MRCNN & COCO & 61.1 & - & - & - & 7.4 & - & - \\
        ProxyDet~\cite{proxy-det} & R50 & CN2~\cite{centernetv2} & LVIS & - & - & 57.0 & - & 19.1 & - & - \\
        CoDet~\cite{codet} & R50 & CN2~\cite{centernetv2} & LVIS, CC3M & - & 39.1 & 57.0 & 14.2 & 20.5 & - & - \\
        Detic~\cite{detic} & Swin-B & CN2~\cite{centernetv2} & LVIS, IN21K & - & - & - & - & 21.5 & 55.2 & - \\
        MMC (text)~\cite{mmc} & R50 & CN2~\cite{centernetv2} & IN-L, LVIS & - & - & - & 16.6 & 23.1 & - & - \\
        3Ways~\cite{3ways} & NF-F0~\cite{nfnet} & FCOS & LVIS & - & 41.5 & - & 16.4 & - & - & - \\
    \midrule
        \belowrulesepcolor{gray!5!}
        \rowcolor{gray!5!}\multicolumn{11}{c}{\textbf{Knowledge Distillation}} \\
        \aboverulesepcolor{gray!5!}
    \midrule
        ViLD~\cite{vild} & R50 & MRCNN & LVIS & 72.2 & 36.6 & 55.6 & 11.8 & 18.2 & - & - \\ 
        CondHead~\cite{condhead} & R50 & ViLD~\cite{vild} & LVIS & 74.6 & 39.1 & 59.1 & 13.2 & 20.4 & - & - \\
        SIC-CADS~\cite{sic-cads} & R50 & Detic~\cite{detic} & LVIS & - & - & - & - & 31.2 & 54.7 & - \\
        BARON~\cite{baron} & R50 & FRCNN & LVIS & 76.0 & 36.2 & 55.7 & 13.6 & 21.0 & - & - \\
        GridCLIP~\cite{grid-clip} & R50 & FCOS & LVIS & 70.9 & 34.7 & 52.2 & - & - & - & - \\
        RKDWTF~\cite{rkdwtf} & R50 & MRCNN & IN21K, LVIS & - & - & 56.6 & - & 22.3 & 42.9 & - \\
        DK-DETR~\cite{dk-detr} & R50 & Def-DETR~\cite{deformable-detr} & LVIS & 71.3 & 39.4 & 54.3 & 12.4 & 17.3 & - & - \\
        DetPro~\cite{detpro} & R50 & MRCNN & LVIS & 74.6 & 34.9 & 53.8 & 12.1 & 18.8 & - & - \\
        CLIPSelf~\cite{clipself} & ViT-L/14 & F-VLM~\cite{f-vlm} & LVIS & - & 40.5 & 63.8 & 19.5 & 31.3 & - & - \\
    \midrule
        \belowrulesepcolor{gray!5!}
        \rowcolor{gray!5!}\multicolumn{11}{c}{\textbf{Transfer Learning}} \\
        \aboverulesepcolor{gray!5!}
    \midrule
        OWL-ViT~\cite{owl-vit} & ViT-B/16 & DETR & O365, VG & - & - & - & - & - & - & 23.6/26.7 \\
        UniDetector~\cite{unidetector} & R50 & FRCNN & COCO, O365, OI & - & - & - & - & - & - & 18.0/19.8 \\
        F-VLM~\cite{f-vlm} & R50 & MRCNN & LVIS & - & 32.5 & 53.1 & 11.9 & 19.2 & - & - \\
        OpenSeeD~\cite{openseed} & Swin-T & Mask DINO~\cite{mask-dino} & COCO, O365 & - & - & - & - & - & - & -/21.8 \\
        Sambor~\cite{sambor} & ViT-B & Cascade R-CNN~\cite{cascade-rcnn} & O365 & - & 48.6 & 66.1 & - & - & - & ${20.9}_{\textcolor{gray}{29.6}}$/${26.3}_{\textcolor{gray}{33.1}}$ \\
    \bottomrule
    \end{tabular}}
    \label{tab:ovd-cdte}
\end{table*}

\begin{table*}
\centering
    \caption{Open-vocabulary semantic segmentation performance on the validation set of ADE20K~\cite{ade20k} (A-847 and A-150), Pascal Context~\cite{pascal-voc} (PC-459 and PC-59), Pascal VOC~\cite{pascal-voc} (PAS-20), Cityscapes~\cite{cityscapes} (CS-19), COCO Stuff~\cite{coco-stuff} (Stuff), and COCO~\cite{mscoco} datasets under the CDTE protocol. MF is MaskFormer~\cite{maskformer}, \emph{c.f.} to~\cref{tab:ovd-mscoco,tab:ovd-lvis} for other notations.}
    \resizebox{\linewidth}{!}{
    \begin{tabular}{lcccccc|cccccccc}
    \toprule
         \multirow{2}{*}{Method} & \multirow{2}{*}{\makecell[c]{Image \\ Backbone}} & \multirow{2}{*}{Segmentor} & \multirow{2}{*}{\makecell[c]{Training Source}} & \multirow{2}{*}{\makecell[c]{Text \\ Encoder}} & \multirow{2}{*}{Prompts} & \multirow{2}{*}{Ensemble} & \multicolumn{8}{c}{mIoU} \\
         & & & & & & & A-847 & A-150 & PC-459 & PC-59 & PAS-20 & CS-19 & Stuff & COCO \\
    \midrule
    \midrule
        \belowrulesepcolor{gray!5!}
        \rowcolor{gray!5!}\multicolumn{14}{c}{\textbf{Region-Aware Training}} \\
        \aboverulesepcolor{gray!5!}
    \midrule
        \multirow{2}{*}{OpenSeg~\cite{openseg}} & \multirow{2}{*}{R101} & \multirow{2}{*}{MF} & \multirow{2}{*}{\makecell[c]{COCO Pan, \\ COCO Cap}} & \multirow{2}{*}{BERT} & \multirow{2}{*}{cat+desc} & \multirow{2}{*}{\XSolidBrush} & \multirow{2}{*}{4.0} & \multirow{2}{*}{15.3} & \multirow{2}{*}{6.5} & \multirow{2}{*}{36.9} & \multirow{2}{*}{60.0} & \multirow{2}{*}{-} & \multirow{2}{*}{-} & \multirow{2}{*}{-} \\
        & & & & & & & & & & & & & & \\
    \noalign{\smallskip}
    \hline
    \noalign{\smallskip}
        \multirow{2}{*}{SLIC~\cite{slic}} & \multirow{2}{*}{ViT-B/16} & \multirow{2}{*}{CAT-Seg~\cite{cat-seg}} & \multirow{2}{*}{\makecell[c]{WebLI~\cite{webli}, \\ COCO Stuff}} & \multirow{2}{*}{CLIP} & \multirow{2}{*}{T (cat)} & \multirow{2}{*}{\XSolidBrush} & \multirow{2}{*}{13.4} & \multirow{2}{*}{36.6} & \multirow{2}{*}{22.0} & \multirow{2}{*}{61.2} & \multirow{2}{*}{95.9} & \multirow{2}{*}{-} & \multirow{2}{*}{-} & \multirow{2}{*}{-} \\ 
        & & & & & & & & & & & & & & \\
    \noalign{\smallskip}
    \hline
    \noalign{\smallskip}
        \multirow{2}{*}{GroupViT~\cite{group-vit}} & \multirow{2}{*}{ViT-S} & \multirow{2}{*}{-} & \multirow{2}{*}{\makecell[c]{CC12M, \\ YFCC14M~\cite{yfcc100m}}} & \multirow{2}{*}{CLIP} & \multirow{2}{*}{T (cat)} & \multirow{2}{*}{\XSolidBrush} & \multirow{2}{*}{-} & \multirow{2}{*}{-} & \multirow{2}{*}{-} & \multirow{2}{*}{22.4} & \multirow{2}{*}{52.3} & \multirow{2}{*}{-} & \multirow{2}{*}{-} & \multirow{2}{*}{-} \\
        & & & & & & & & & & & & & & \\
    \noalign{\smallskip}
    \hline
    \noalign{\smallskip}
        ViL-Seg~\cite{vil-seg} & ViT-B/16 & - & CC12M & CLIP & T (cat) & \XSolidBrush & - & - & - & 16.3 & 34.4 & - & 16.4 & - \\
        SegCLIP~\cite{segclip} & ViT & - & COCO Cap, CC3M & CLIP & T (cat) & \XSolidBrush & - & - & - & 24.7 & 52.6 & - & - & 26.5 \\
        OVSegmentor~\cite{ovsegmentor} & ViT-B & - & CC4M~\cite{ovsegmentor} & BERT & T (cat) & \XSolidBrush & - & 5.6 & - & 20.4 & 53.8 & - & - & 25.1 \\
    \noalign{\smallskip}
    \hline
    \noalign{\smallskip}
        \multirow{2}{*}{PACL~\cite{pacl}} & \multirow{2}{*}{ViT-B/16} & \multirow{2}{*}{-} & \multirow{2}{*}{\makecell[c]{CC3M, CC12M, \\ YFCC15M~\cite{yfcc100m}}} & \multirow{2}{*}{CLIP} & \multirow{2}{*}{T (cat)} & \multirow{2}{*}{\XSolidBrush} & \multirow{2}{*}{-} & \multirow{2}{*}{31.4} & \multirow{2}{*}{-} & \multirow{2}{*}{50.1} & \multirow{2}{*}{72.3} & \multirow{2}{*}{-} & \multirow{2}{*}{38.8} & \multirow{2}{*}{-} \\
        & & & & & & & & & & & & & & \\
    \noalign{\smallskip}
    \hline
    \noalign{\smallskip}
        TCL~\cite{tcl} & ViT-B/16 & - & CC3M, CC12M & CLIP & T (cat) & \XSolidBrush & - & 17.1 & - & 33.9 & 83.2 & 24.0 & 22.4 & 31.6 \\
        SimSeg~\cite{simseg} & ViT-B/16 & - & CC3M, CC12M & CLIP & T (cat) & \XSolidBrush & - & - & - & 26.2 & 57.4 & - & 29.7 & - \\
    \midrule
        \belowrulesepcolor{gray!5!}
        \rowcolor{gray!5!}\multicolumn{15}{c}{\textbf{Knowledge Distillation}} \\
        \aboverulesepcolor{gray!5!}
    \midrule
        \multirow{2}{*}{GKC~\cite{gkc}} & \multirow{2}{*}{R50} & \multirow{2}{*}{MF} & \multirow{2}{*}{\makecell[c]{COCO Pan, \\ COCO Cap}} & \multirow{2}{*}{CLIP} & \multirow{2}{*}{T (cat+desc)} & \multirow{2}{*}{\XSolidBrush} & \multirow{2}{*}{3.2} & \multirow{2}{*}{17.5} & \multirow{2}{*}{6.5} & \multirow{2}{*}{41.9} & \multirow{2}{*}{78.7} & \multirow{2}{*}{34.3} & \multirow{2}{*}{-} & \multirow{2}{*}{-} \\
        & & & & & & & & & & & & & & \\
    \noalign{\smallskip}
    \hline
    \noalign{\smallskip}
        \multirow{3}{*}{SAM-CLIP~\cite{sam-clip}} & \multirow{3}{*}{ViT-B} & \multirow{3}{*}{SAM~\cite{sam}} & \multirow{3}{*}{\makecell[c]{CC3M, CC12M, \\ YFCC15M~\cite{yfcc100m}, \\ IN21K, SA-1B~\cite{sam}}} & \multirow{3}{*}{CLIP} & \multirow{3}{*}{T (cat)} & \multirow{3}{*}{\XSolidBrush} & \multirow{3}{*}{-} & \multirow{3}{*}{-} & \multirow{3}{*}{-} & \multirow{3}{*}{29.2} & \multirow{3}{*}{60.6} & \multirow{3}{*}{17.1} & \multirow{3}{*}{31.5} & \multirow{3}{*}{-} \\
        & & & & & & & & & & & & & & \\
        & & & & & & & & & & & & & & \\
    \noalign{\smallskip}
    \hline
    \noalign{\smallskip}
        ZeroSeg~\cite{zeroseg} & ViT & - & IN1K~\cite{ilsvrc} & CLIP & T (cat) & \XSolidBrush & - & - & - & 20.4 & 40.8 & - & - & 20.2 \\
    \midrule
        \belowrulesepcolor{gray!5!}
        \rowcolor{gray!5!}\multicolumn{15}{c}{\textbf{Transfer Learning}} \\
        \aboverulesepcolor{gray!5!}
    \midrule
        LSeg+~\cite{openseg} & R101 & SRB~\cite{lseg} & COCO Pan & CLIP & T (cat) & \XSolidBrush & 2.5 & 13.0 & 5.2 & 36.0 & 59.0 & - & - & - \\
    \noalign{\smallskip}
    \hline
    \noalign{\smallskip}
        \multirow{2}{*}{CEL~\cite{cel}} & \multirow{2}{*}{R50} & \multirow{2}{*}{MF} & \multirow{2}{*}{\makecell[c]{COCO Cap, \\ COCO Stuff}} & \multirow{2}{*}{CLIP} & \multirow{2}{*}{T (cat)} & \multirow{2}{*}{\Checkmark} & \multirow{2}{*}{7.2} & \multirow{2}{*}{20.5} & \multirow{2}{*}{9.6} & \multirow{2}{*}{49.6} & \multirow{2}{*}{86.7} & \multirow{2}{*}{-} & \multirow{2}{*}{-} & \multirow{2}{*}{-} \\ 
        & & & & & & & & & & & & & & \\
    \noalign{\smallskip}
    \hline
    \noalign{\smallskip}
        ZSSeg~\cite{zsseg} & R101 & MF & COCO Stuff & CLIP & L (cat) & \Checkmark & 7.0 & 20.5 & - & 47.7 & - & 34.5 & - & - \\
        MaskCLIP~\cite{maskclip} & R101 & DL~\cite{deeplab} & \XSolidBrush & CLIP & T (cat) & \XSolidBrush & - & - & - & 25.5 & - & - & 14.6 & - \\
        CLIP-DINOiser~\cite{clip-denoiser} & ViT-B/16 & - & Pascal VOC & CLIP & T (cat) & \XSolidBrush & - & 20.0 & - & 35.9 & 80.2 & 31.7 & - & - \\
        MVP-SEG~\cite{mvp-seg} & R50 & DL~\cite{deeplab} & COCO Stuff & CLIP & L (cat) & \XSolidBrush & - & - & - & 38.7 & - & - & - & - \\
        ReCo~\cite{reco} & - & - & - & CLIP & T (cat) & \XSolidBrush & - & - & - & - & - & 19.3 & 26.3 & - \\
    \noalign{\smallskip}
    \hline
    \noalign{\smallskip}
        \multirow{2}{*}{OVDiff~\cite{ovdiff}} & \multirow{2}{*}{UNet~\cite{unet}} & \multirow{2}{*}{-} & \multirow{2}{*}{\makecell[c]{CLIP~\cite{clip}, \\ Stable Diffusion~\cite{stable-diffusion}}} & \multirow{2}{*}{\XSolidBrush} & \multirow{2}{*}{T (cat)} & \multirow{2}{*}{\XSolidBrush} & \multirow{2}{*}{-} & \multirow{2}{*}{-} & \multirow{2}{*}{-} & \multirow{2}{*}{30.1} & \multirow{2}{*}{67.1} & \multirow{2}{*}{-} & \multirow{2}{*}{-} & \multirow{2}{*}{34.8} \\
        & & & & & & & & & & & & & & \\
    \noalign{\smallskip}
    \hline
    \noalign{\smallskip}
        FOSSIL~\cite{fossil} & ViT-L/14 & - & COCO Cap & CLIP & T (cat) & \XSolidBrush & - & - & - & 35.8 & - & 23.2 & 24.8 & - \\
        POMP~\cite{pomp} & R101 & MF & COCO Stuff & CLIP & L (cat) & \XSolidBrush & - & 20.7 & - & 51.1 & - & - & - & - \\
        AttrSeg~\cite{attrseg} & R101 & - & COCO Stuff & CLIP & desc & \XSolidBrush & - & - & - & 56.3 & 91.6 & - & - & - \\ 
        PnP-OVSS~\cite{pnp-ovss} & ViT-L/16 & BLIP~\cite{blip} & COCO Cap & BERT & T (cat) & \XSolidBrush & - & 23.2 & - & 41.9 & 55.7 & - & 32.6 & 33.8 \\ 
        SCAN~\cite{scan} & Swin-B & M2F & COCO Stuff & CLIP & T (cat) & \XSolidBrush & 10.8 & 30.8 & 13.2 & 58.4 & 97.0 & - & - & - \\
        TagAlign~\cite{tagalign} & ViT-B/16 & - & CC12M & CLIP & T (cat) & \XSolidBrush & - & 17.3 & - & 37.6 & 87.9 & 27.5 & - & 33.3 \\ 
        Self-Seg~\cite{self-seg} & ViT-L & X-Dec~\cite{x-decoder} & COCO Cap & - & - & \XSolidBrush & 6.4 & - & - & - & - & 41.1 & - & - \\ 
    \noalign{\smallskip}
    \hline
    \noalign{\smallskip}
        OVSeg~\cite{ovseg} & R101c~\cite{deeplab} & MF & \makecell[c]{COCO Stuff, \\ COCO Cap} & CLIP & T (cat) & \Checkmark & 7.1 & 24.8 & 11.0 & 53.3 & 92.6 & - & - & - \\
    \noalign{\smallskip}
    \hline
    \noalign{\smallskip}
        CAT-Seg~\cite{cat-seg} & Swin-B & - & COCO Stuff & CLIP & T (cat) & \XSolidBrush & 10.8 & 31.5 & 20.4 & 62.0 & 96.6 & - & - & - \\
        SED~\cite{sed} & ConvNext-B~\cite{convnext} & - & COCO Stuff & CLIP & T (cat) & \XSolidBrush & 11.4 & 31.6 & 18.6 & 57.3 & 94.4 & - & - & - \\
        MAFT~\cite{maft} & ViT-B/16 & FreeSeg~\cite{freeseg} & COCO Stuff & CLIP & T (cat) & \XSolidBrush & 10.1 & 29.1 & 12.8 & 53.5 & 90.0 & - & - & - \\
        SAN~\cite{san} & ViT-B/16 & - & COCO Stuff & CLIP & T (cat) & \XSolidBrush & 10.1 & 27.5 & 12.6 & 53.8 & 94.0 & - & - & - \\
        CaR~\cite{car} & ViT-B/16 & - & \XSolidBrush & CLIP & T (cat) & \XSolidBrush & - & - & - & 30.5 & 67.6 & - & - & 36.6 \\
    \bottomrule
    \end{tabular}}
    \label{tab:ovs-ovss-cdte}
\end{table*}

\begin{table*}[t!]
    \centering
    \caption{Open-vocabulary semantic segmentation performance under generalized evaluation protocol. HM is the hamonic mean (hIoU), \emph{c.f.} to~\cref{tab:ovd-mscoco,tab:ovd-lvis,tab:ovs-ovss-cdte} for other notations.} 
    \resizebox{\linewidth}{!}{
    \begin{tabular}{lcccccc|c|c}
    \toprule
         \multirow{2}{*}{Method} & \multirow{2}{*}{\makecell[c]{Image \\ Backbone}} & \multirow{2}{*}{Segmentor} & \multirow{2}{*}{\makecell[c]{Image-Text \\ Pairs}} & \multirow{2}{*}{\makecell[c]{Text \\ Encoder}} & \multirow{2}{*}{Prompts} & \multirow{2}{*}{Ensemble} & Pascal VOC & COCO Stuff \\
         & & & & & & & mIoU(N/B/HM) & mIoU(N/B/HM) \\
    \midrule
    \midrule
        CEL~\cite{cel} & R50 & MF & COCO Cap & CLIP & T (cat) & \Checkmark & 74.8/88.5/81.1 & 42.0/38.6/40.2 \\
        ZegFormer~\cite{zegformer} & R101 & MF & \XSolidBrush & CLIP & T (cat) & \Checkmark & 63.6/86.4/73.3 & 33.2/36.6/34.8 \\
        ZSSeg~\cite{zsseg} & R101 & MF & \XSolidBrush & CLIP & L (cat) & \Checkmark & 72.5/83.5/77.5 & 36.3/39.3/37.8 \\
        MVP-SEG+~\cite{mvp-seg} & R50 & DL~\cite{deeplab} & \XSolidBrush & CLIP & L (cat) & \XSolidBrush & 87.4/89.0/88.2 & 55.8/38.3/45.5 \\
        POMP~\cite{pomp} & R101 & MF & \XSolidBrush & CLIP & L (cat) & \XSolidBrush & 76.8/93.6/84.4 & 38.2/39.9/39.1 \\
        MAFT~\cite{maft} & ViT-B/16 & FreeSeg~\cite{freeseg} & \XSolidBrush & CLIP & T (cat) & \XSolidBrush & 81.8/91.4/86.3 & 50.4/43.3/46.5 \\
        ZegCLIP~\cite{zegclip} & ViT-B/16 & SegViT~\cite{segvit} & \XSolidBrush & CLIP & T (cat) & \XSolidBrush & 77.8/91.9/84.3 & 41.4/40.2/40.8 \\
        TagCLIP~\cite{tagclip} & ViT-B/16 & SegViT~\cite{segvit} & \XSolidBrush & CLIP & T (cat) & \XSolidBrush & 85.2/93.5/89.2 & 43.1/40.7/41.9 \\ 
    \bottomrule
    \end{tabular}}
    \label{tab:ovs-ovss-voc-stuff-ade20k}
\end{table*}

\begin{table*}
\centering
    \caption{Open-vocabulary instance segmentation performance on COCO~\cite{mscoco} and OpenImages~\cite{open-images-v4} datasets under the gOVE protocol. The metric is mask AP. M2F is Mask2Former~\cite{mask2former}, \emph{c.f.} to~\cref{tab:ovd-mscoco,tab:ovd-lvis} for other notations.}
    \resizebox{\linewidth}{!}{
    \begin{tabular}{lcccccc|ccc|ccc}
    \toprule
         \multirow{2}{*}{Method} & \multirow{2}{*}{\makecell[c]{Image \\ Backbone}} & \multirow{2}{*}{Segmentor} & \multirow{2}{*}{\makecell[c]{Image-Text \\ pairs}} & \multirow{2}{*}{\makecell[c]{Text \\ Encoder}} & \multirow{2}{*}{Prompts} & \multirow{2}{*}{Ensemble} & \multicolumn{3}{c}{COCO} \vline & \multicolumn{3}{c}{OpenImages} \\
         & & & & & & & $\text{AP}_{50}^{N}$ & $\text{AP}_{50}^{B}$ & $\text{AP}_{50}$ & $\text{AP}_{50}^{N}$ & $\text{AP}_{50}^{B}$ & $\text{AP}_{50}$ \\
    \midrule
    \midrule
        CGG~\cite{cgg} & R50 & M2F & COCO Cap & BERT & \XSolidBrush & \XSolidBrush & 28.4 & 46.0 & 41.4 & - & - & - \\
        $\text{D}^{2}\text{Zero}$~\cite{d2zero} & R50 & M2F & \XSolidBrush & CLIP & T (cat) & \XSolidBrush & 15.8 & 54.1 & 24.5 & - & - & - \\
        XPM~\cite{xpm} & R50 & MRCNN & CC3M & BERT & \XSolidBrush & \XSolidBrush & 21.6 & 41.5 & 36.3 & 22.7 & 49.8 & 40.7 \\
        Mask-free OVIS & R50 & MRCNN & COCO, OI & ALBEF~\cite{albef} & \XSolidBrush & \XSolidBrush & 25.0 & - & - & 25.8 & - & - \\
    \bottomrule
    \end{tabular}}
    \label{tab:ovs-ovis-coco-open-images}
\end{table*}

\begin{table*}
\centering
    \caption{Open-vocabulary panoptic segmentation performance on COCO Panoptic~\cite{panoptic-segmentation} and ADE20k~\cite{ade20k} dataset. $\text{PQ}^{st}$ and $\text{PQ}^{th}$ represent PQ for stuff and thing classes, respectively. For other notations, \emph{c.f.} to~\cref{tab:ovd-mscoco,tab:ovd-lvis,tab:ovs-ovis-coco-open-images}.}
    \resizebox{\linewidth}{!}{
    \begin{tabular}{lccccc|cccccc|ccccc}
    \toprule
         \multirow{2}{*}{Method} & \multirow{2}{*}{\makecell[c]{Image \\ Backbone}} & \multirow{2}{*}{Segmentor} & \multirow{2}{*}{\makecell[c]{Text \\ Encoder}} & \multirow{2}{*}{Prompts} & \multirow{2}{*}{Ensemble} & \multicolumn{6}{c}{COCO Panoptic} \vline & \multicolumn{5}{c}{ADE20K} \\ 
         & & & & & & $\text{PQ}^{B}$ & $\text{SQ}^{B}$ & $\text{RQ}^{B}$ & $\text{PQ}^{N}$ & $\text{SQ}^{N}$ & $\text{RQ}^{N}$ & PQ & $\text{PQ}^{th}$ & $\text{PQ}^{st}$ & SQ & RQ \\
    \midrule
    \midrule
        PADing~\cite{pading} & R50 & M2F & CLIP & T (cat) & \XSolidBrush & 41.5 & 80.6 & 49.7 & 15.3 & 72.8 & 18.4 & - & - & - & - & - \\
        FreeSeg~\cite{freeseg} & R101 & M2F & CLIP & L (cat) & \XSolidBrush & 31.4 & 78.3 & 38.9 & 29.8 & 79.2 & 37.6 & - & - & - & - & - \\
        MaskCLIP~\cite{maskclip2} & R50 & M2F & CLIP & cat & \XSolidBrush & - & - & - & - & - & - & 15.1 & 13.5 & 18.3 & 70.5 & 19.2 \\
        OPSNet~\cite{opsnet} & R50 & M2F & CLIP & cat & \XSolidBrush & - & - & - & - & - & - & 17.7 & 15.6 & 21.9 & 54.9 & 21.6 \\
    \bottomrule
    \end{tabular}}
    \label{tab:ovs-ovps-coco-ade20k}
\end{table*}

\begin{table*}
\centering
    \caption{Open-vocabulary panoptic segmentation performance under the CDTE protocol. For notations, \emph{c.f.} to~\cref{tab:ovs-ovps-coco-ade20k}.}
    \resizebox{\linewidth}{!}{
    \begin{tabular}{lcccccc|ccc|ccc|ccc}
    \toprule
         \multirow{2}{*}{Method} & \multirow{2}{*}{\makecell[c]{Image \\ Backbone}} & \multirow{2}{*}{Segmentor} & \multirow{2}{*}{\makecell[c]{Text \\ Encoder}} & \multirow{2}{*}{Prompts} & \multirow{2}{*}{\makecell[c]{Training Source}} & \multirow{2}{*}{Ensemble} & \multicolumn{3}{c}{COCO Pan} \vline & \multicolumn{3}{c}{ADE20K} \vline & \multicolumn{3}{c}{Cityscapes} \\ 
         & & & & & & & PQ & SQ & RQ & PQ & SQ & RQ & PQ & SQ & RQ \\
    \midrule
    \midrule
        Uni-OVSeg~\cite{uni-ovseg} & ConvNext-L~\cite{convnext} & M2F & CLIP & T (cat) & \emph{c.f.} to~\cite{uni-ovseg} & \XSolidBrush & 18.0 & 72.6 & 24.3 & 14.1 & 66.1 & 19.0 & 17.5 & 65.2 & 23.5 \\
    \noalign{\smallskip}
    \hline
    \noalign{\smallskip}
        X-Decoder~\cite{x-decoder} & DaViT-B~\cite{davit} & M2F & CLIP & T (cat) & \makecell[c]{CC3M, SBU~\cite{sbu}, \\ COCO Cap, VG} & \XSolidBrush & - & - & - & 21.1 & - & - & 39.5 & - & - \\
    \noalign{\smallskip}
    \hline
    \noalign{\smallskip}
        APE~\cite{ape} & ViT-L & - & CLIP & T (cat+desc) & \emph{c.f.} to~\cite{ape} & \XSolidBrush & - & - & - & 26.1 & - & - & 32.8 & - & - \\
        FC-CLIP~\cite{fc-clip} & ConvNext-L~\cite{convnext} & M2F & CLIP & T (cat) & COCO Pan or ADE20K & \Checkmark & 27.0 & 78.0 & 32.9 & 26.8 & - & - & 44.0 & 75.4 & 53.6 \\
        PosSAM~\cite{possam} & ViT-B & M2F & CLIP & T (cat) & COCO Pan or ADE20K & \XSolidBrush & 25.1 & 80.1 & 30.4 & 26.5 & 74.7 & 32.0 & - & - & - \\
        MasQCLIP~\cite{masqclip} & R50 & M2F & CLIP & cat & COCO Pan & \XSolidBrush & - & - & - & 23.3 & - & - & - & - & - \\
        OMG-Seg~\cite{omg-seg} & ConvNext-L~\cite{convnext} & M2F & CLIP & T (cat) & \emph{c.f.} to ~\cite{omg-seg} & \XSolidBrush & 53.8 & - & - & - & - & - & 65.7 & - & - \\
        ODISE~\cite{odise} & UNet~\cite{unet} & M2F & CLIP & T (cat+desc) & COCO Pan, COCO Cap & \Checkmark & - & - & - & 23.4 & - & - & - & - & - \\
        HIPIE~\cite{hipie} & R50 & \makecell[c]{Def-DETR~\cite{deformable-detr}, \\ MaskDINO~\cite{mask-dino}} & BERT & cat & \emph{c.f.} to~\cite{hipie} & \XSolidBrush & - & - & - & 18.4 & - & - & - & - & - \\
    \bottomrule
    \end{tabular}}
    \label{tab:ovs-ovps-cdte}
\end{table*}